%% file: main.tex
\documentclass[runningheads]{llncs}

\usepackage{eccv}

\usepackage{eccvabbrv}

\usepackage{graphicx}
\usepackage{booktabs}

\usepackage[accsupp]{axessibility}  

\usepackage[pagebackref]{hyperref}

\usepackage{orcidlink}

\usepackage{amsmath}
\usepackage{amssymb}
\usepackage{graphicx}
\usepackage{booktabs}
\usepackage[table,xcdraw]{xcolor}
\usepackage{caption}
\usepackage{multirow}
\usepackage{multicol}
\usepackage{array}
\usepackage{tabularx}
\usepackage{floatrow}
\usepackage{subcaption}
\usepackage{wrapfig}

\definecolor{Gray}{HTML}{EFEFEF}
\newcommand{\tok}{{WinTok}}

\begin{document}

\title{WinTok: A Win-Win Hybrid Tokenizer via Decomposing Visual Understanding and Generation with Transferable Tokens} 

\titlerunning{WinTok}


\author{Yiwei Guo\inst{\ast1} \and
Shaobin Zhuang\inst{\ast3} \and
Zhipeng Huang\inst{2} \and
Canmiao Fu\inst{2} \\
Chen Li\inst{2} \and
Jing LYU\inst{2} \and
Yali Wang\inst{\dagger1}}

\authorrunning{Y. Guo et al.}

\institute{
Shenzhen Institutes of Advanced Technology, Chinese Academy of Sciences \and
WeChat Vision, Tencent Inc. \and
Shanghai Jiao Tong University
}

\maketitle
\def\thefootnote{$\ast$}\footnotetext{Interns at WeChat Vision, Tencent Inc. $\quad$ $\dagger$ Corresponding Author.}
\def\thefootnote{\arabic{footnote}}

\input{sec/0_abstract}
\input{sec/1_intro}
\input{sec/2_related}
\input{sec/3_method}
\input{sec/4_exp}
\input{sec/5_conclusion}
\input{sec/X_suppl}



%
%
\bibliographystyle{splncs04}
\bibliography{main}
\end{document}

%% file: sec/0_abstract.tex
\begin{abstract}
\label{sec:abs}

Building a unified visual tokenizer is essential for bridging the gap between visual understanding and generation. 
Yet existing approaches struggle with the inherent conflict between these tasks, 
as a single token space is forced to support both high-level semantic abstraction and low-level pixel reconstruction.
We propose \textbf{WinTok}, a concise hybrid tokenizer that achieves a win-win performance by 
explicitly decoupling the two objectives. WinTok supplements pixel tokens with a set of learnable semantic tokens, 
effectively mitigating cross-task interference without incurring the computational overhead of dual tokenizers.
To further enhance understanding capability, we introduce an asymmetric token distillation mechanism: 
the semantic tokens are guided by pretrained semantic embeddings from any visual foundation model, 
enabling them to inherit strong discriminative power while maintaining flexibility.
Across \textbf{10} challenging benchmarks, WinTok delivers consistent improvements in reconstruction, understanding, and generation. 
Trained on only 50M open-source data, WinTok surpasses the strong baseline UniTok by \textbf{11.2\%} in classification accuracy 
and achieves a competitive reconstruction rFID of \textbf{0.41}, despite using substantially less training data. Code is released at https://github.com/markywg/WinTok.
\keywords{Unified tokenizer \and Hybrid encoding \and Transferable tokens}
\end{abstract}

%% file: sec/1_intro.tex
\section{Introduction}
\label{sec:intro}


The emergence of large language models (LLMs) has fundamentally reshaped the landscape of natural language processing by introducing a unified next-token prediction paradigm \cite{brown2020gpt3,touvron2023llama,bai2023qwen}.
Building upon this principle, recent efforts have extended the unified autoregressive framework to jointly model visual understanding and generation within a single multimodal architecture \cite{hurst2024gpt4o,team2023gemini,team2024chameleon,sun2024emu,zhou2025transfusion,xie2025show}.
However, these unified models face a persistent dilemma: 
Despite these advances, unified multimodal models confront an inherent tension: visual understanding and visual generation require tokens with distinct granularities and representational forms. Visual understanding favors high-level continuous tokens that capture abstract and semantically rich representations for image comprehension 
\cite{wang2024qwen2vl,liu2023visual,liu2024improved}.
In contrast, visual generation relies on low-level discrete tokens to enable precise and high-fidelity pixel synthesis \cite{esser2021taming,sun2024llamagen,wu2024liquid}.

\begin{figure}[!t]
  \centering
  \includegraphics[width=\linewidth]{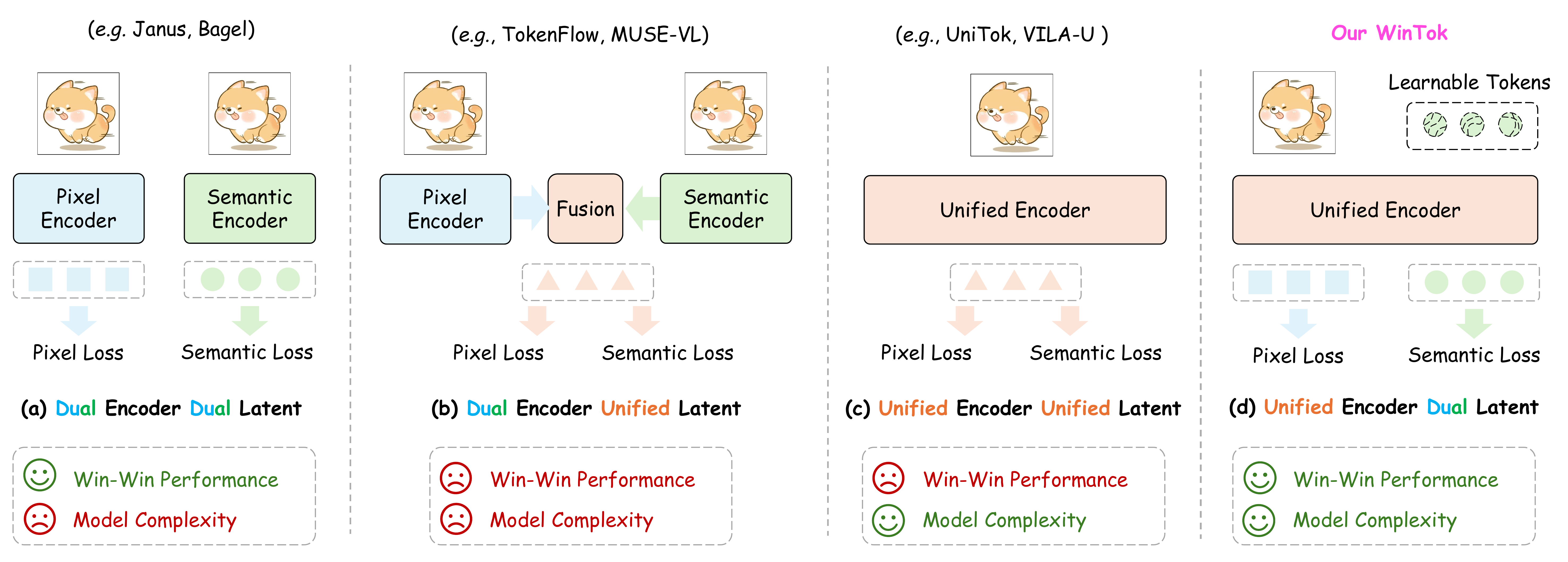}

   \caption{\textbf{Comparsion of different tokenization paradigms.}(a) Dual encoders with separate representations obtain plausible performance at the cost of model complexity.
    (b-c) Previous unified tokenizers face representation conflict due to contradictory optimization objectives, thus leading to performance trade-offs.
    (d) Our WinTok decomposes visual understanding and generation with learnable tokens, achieving a win-win performance with reduced complexity.
   }
   \label{fig:motivation}
\end{figure}

To reconcile these divergent requirements, 
most existing unified multi-modal models typically employ a separate visual tokenizer for each task \cite{wu2025janus,chen2025januspro,zhuang2025vargpt},
as shown in \cref{fig:motivation} (a).
Semantic encoders \cite{radford2021clip,zhai2023sigmoid,tschannen2025siglip} extract continuous tokens for understanding,
while pixel encoders \cite{van2017vqvae,yu2022vqgan,esser2021taming} generate discrete tokens for generation.
However,
this would introduce significant complexity,
without achieving fundamental model unification.
To tackle this core issue,
recent efforts have been made to construct a unified tokenizer.
One type is the dual encoder with fusion (\eg, shared mapping or MLP) \cite{qu2025tokenflow,xie2025muse},
as shown in \cref{fig:motivation} (b), which inherit the complexities of dual architectures.
Alternatively,
encoder unification is preferable \cite{ma2025unitok,zhao2025qlip},
as shown in \cref{fig:motivation} (c). 
But unified encoders force a single set of visual tokens to handle both high-level semantic abstraction and low-level pixel reconstruction,
leading to performance trade-offs between understanding and generation due to optimization conflict \cite{song2025dualtoken,wu2025harmonizing,lin2025toklip,tang2025unilip,yue2025uniflow}.

\setlength{\textfloatsep}{8pt}

To address this conflict, 
we introduce WinTok, 
which is a hybrid tokenizer that achieves a win-win performance,
by decomposing visual understanding and generation with transferable tokens.
As illustrated in \cref{fig:motivation} (d),
WinTok comprises two types of tokens,
including pixel and semantic tokens.
On one hand,
the unified encoder receives the input image to generate the pixel tokens.
Since these pixel tokens often contain rich local details,
they are suitable for visual generation.
On the other hand,
we introduce a set of learnable tokens as extra input to the unified encoder to obtain semantic tokens.
To effectively equip these learnable tokens with high-level semantics,
we introduce an asymmetric token distillation paradigm in our WinTok. 
Specifically,
we leverage any visual foundation model \cite{radford2021clip, zhai2023sigmoid, tschannen2025siglip} as the semantic teacher to extract well-pretrained semantic tokens of the input image.
Then,
we treat them as semantic supervision to guide the optimization of learnable tokens.
We refer to this as asymmetric token distillation,
since semantic knowledge is transferred from the input image to learnable tokens.
Through this token-level knowledge transfer paradigm, the learnable tokens progressively inherit and internalize the representational capabilities of foundation models, thereby acquiring enhanced capacity to model high-level visual semantics.

The interaction between the two types of tokens in the unified encoder enables them to 
adaptively integrate complementary contextual information, 
optimizing local details for generation and global semantics for understanding. 
Extensive experiments demonstrate that WinTok outperforms pioneering unified tokenizers 
across 10 mainstream benchmarks for visual reconstruction, understanding, and generation.
Specifically,
on ImageNet-1K \cite{deng2009imagenet} validation set,
WinTok achieves an \textbf{82.0\%} Top-1 accuracy, surpassing the strong counterpart UniTok \cite{ma2025unitok}
by \textbf{11.2\%}, while maintaining a competitive reconstruction quality with an rFID of \textbf{0.41}
using significantly fewer data and a reduced codebook size.
Moreover, WinTok demonstrates leading performance on downstream multimodal comprehension and visual generation,
outperforming UniTok on POPE \cite{li2023pope} by 3.3\% and delivering a better performance on GenEval (0.76 vs. 0.59).
These results indicate that WinTok effectively balances the needs of both visual understanding and generation tasks.
We believe that WinTok offers a promising direction for future research in unified multimodal modeling.

%% file: sec/2_related.tex
\section{Related Work}
\label{sec:related}

\subsection{Visual Tokenizer for Understanding}
To extend the capability of large language models (LLMs) to comprehend visual content,
numerous multimodal large language models (MLLMs) have been proposed \cite{liu2023visual,liu2024improved,li2023blip2,wang2024qwen2vl,chen2024internvl}.
These methods typically employ a language-aligned semantic visual tokenizer \cite{radford2021clip,zhai2023sigmoid,tschannen2025siglip}
to extract continuous visual tokens that encapsulate high-level semantic information from images.
However, these tokenizers are primarily designed for visual understanding tasks,
and thus may not effectively capture the fine-grained details necessary for faithful image reconstruction or generation \cite{song2025dualtoken,tang2025unilip}.

\subsection{Visual Tokenizer for Generation}
In the visual generation domain, mainstream approaches \cite{rombach2022sd,peebles2023dit,esser2021taming,chang2022maskgit,tian2024visual} utilize reconstruction-oriented visual tokenizers \cite{kingma2013vae,van2017vqvae,yu2022vqgan}
to map images into a compact latent space, which reduces computational overhead and modeling complexity.
For example, diffusion models \cite{rombach2022sd,peebles2023dit,ma2024sit} employ VAE \cite{kingma2013vae,kingma2019introduction}
to encode images into continuous tokens while autoregressive models \cite{chang2022maskgit,tian2024visual,sun2024llamagen}
leverage VQVAE \cite{van2017vqvae,yu2022vqgan} to convert images into discrete tokens.
More recently, some works \cite{zheng2025vfmtok,chen2025aligning,bi2025vfmvae,zheng2025rae} have explored to transfer vision foundation models (VFMs) \cite{oquab2023dinov2,radford2021clip,tschannen2025siglip}
as visual tokenizers for generative tasks.
Others introduce 1D-tokenizers \cite{yu2024image, li2024imagefolder, bachmann2025flextok} for better token efficiency, or multi-codebook quantization \cite{jia2025mgvq, bai2024factorized} for better codebook learning.
Nonetheless, these reconstruction-oriented tokenizers may not effectively capture high-level semantic information essential for understanding tasks \cite{qu2025tokenflow,wu2025harmonizing}.

\subsection{Unified Visual Tokenizer}
Early unified multimodal models (UMMs) \cite{wu2025janus,chen2025januspro,deng2025bagel} typically adopt separate visual tokenizers for understanding and generation tasks,
which leads to increased model complexity and training costs.
To bridge this gap, recent efforts have focused on unified visual tokenizers. Some methods combine semantic and pixel encoders via late-fusion, targeting either distinct codebook learning \cite{qu2025tokenflow,xie2025muse} or hierarchical feature integration \cite{chen2025semhitok,lin2025toklip}; however, this paradigm remains cumbersome. 
Alternatively, single-encoder approaches yield unified representations through explicit semantic alignment \cite{wu2024vila,zhao2025qlip}, shared latent optimization \cite{ma2025unitok,tang2025unilip,li2025manzano}, or dual-stream balancing \cite{song2025dualtoken,yue2025uniflow}. While structurally simpler, they often struggle to balance conflicting training objectives. Though contemporary work VQRAE \cite{du2025vqrae} addresses this via a hybrid tokenizer, it relies on an elaborate two-stage training scheme. In contrast, \tok~mitigates these obstacles by introducing transferable tokens to decompose the conflicting objectives, enabling a harmonious balance between visual understanding and generation.

%% file: sec/3_method.tex
\section{Method}
In this section, 
we first provide the preliminary knowledge on typical tokenizers used for visual understanding and generation.
Motivated by representation conflict between these tasks,
we then elaborate on our WinTok framework, 
incorporating asymmetric token distillation.

\subsection{Preliminary}

\noindent\textbf{Semantic Tokenizers for Visual Understanding.}
Semantic tokenizers convert raw pixels into a compact sequence of high-level semantic tokens for visual understanding. 
The scaling of data and models in recent foundation models has demonstrated remarkable discriminative power 
by utilizing vision transformers \cite{dosovitskiy2021vit} alongside various 
self-supervised learning and multimodal alignment strategies \cite{he2022mae,oquab2023dinov2,radford2021clip,zhai2023sigmoid,tschannen2025siglip,gui2024survey}.
These models are considered effective semantic tokenizers 
because they leverage continuous semantic tokens to address downstream understanding tasks \cite{liu2023visual,liu2024improved,alayrac2022flamingo}.


\noindent\textbf{Pixel Tokenizers for Visual Generation.}
To facilitate auto-regressive visual generation, 
several pixel tokenizers have been developed using Vector-Quantized Variational Autoencoders (VQVAEs) \cite{van2017vqvae,yu2022vqgan,esser2021taming}.
These tokenizers generally consist of an encoder, a quantizer, and a decoder. 
Given an input image, the encoder converts it into a set of continuous latent tokens. 
The quantizer then transforms these continuous tokens into discrete visual tokens by 
identifying their nearest embeddings within a learnable codebook. 
Finally, the decoder reconstructs the image from the quantized tokens.
By combining pixel reconstruction loss with codebook learning loss \cite{yu2022vqgan,sun2024llamagen}, 
these tokenizers achieve high-fidelity reconstruction and generation.

\begin{wraptable}{r}{6.5cm}
  \caption{\textbf{Comparison of adapting different visual tokenizers for understanding and reconstruction.}
  }
  \label{tab:rep_conflict}
  \centering
  \small
  \begin{tabular}{@{}llcc@{}}
    \toprule
    \textbf{Visual Tokenizer} & \textbf{Type} & \textbf{rFID $\downarrow$} & \textbf{Acc.$\uparrow$} \\
    \midrule
    SigLIP2 \cite{tschannen2025siglip} & Semantic & 0.82 & 81.7 \\
    WeTok \cite{zhuang2025wetok} & Pixel & 0.64  & 18.3 \\
     \midrule
    WinTok & Hybrid & \textbf{0.41}  & \textbf{82.0} \\
    \bottomrule
  \end{tabular}
\end{wraptable}

\noindent\textbf{Conflict between Understanding and Generation.}
As discussed above,
visual understanding and generation typically depend on different tokenizers with different visual granularities and formulations to process images.
Hence,
it would be inappropriate to use pixel tokenizer for understanding and use semantic tokenizer for generation.
We further conduct an experiment on ImageNet \cite{deng2009imagenet} to validate this observation using two state-of-the-art tokenizers,
\ie,
SigLIP2 \cite{tschannen2025siglip} as the semantic tokenizer,
and
WeTok \cite{zhuang2025wetok} as the pixel tokenizer.
As shown in \cref{tab:rep_conflict},
SigLIP2 exhibits strong capability in understanding (\eg, classification accuracy) but deteriorates in reconstruction (\eg, rFID).
In contrast, 
WeTok excels in reconstruction quality but shows significant declines in understanding performance.
These findings underscore that adapting existing visual tokenizers for both understanding and generation is suboptimal 
due to the conflicting goals.
This fundamental conflict hinders the development of a unified tokenizer \cite{qu2025tokenflow,ma2025unitok}.

\subsection{WinTok}

Based on the above discussion,
a natural question arises:
\textit{Is it feasible to build a unified tokenizer while decomposing visual understanding and generation?}
To answer this question,
we introduce WinTok, a hybrid tokenizer incorporated with transferable tokens to decompose visual understanding and generation
and achieve a win-win performance.

\begin{figure}[!t]
    \centering
    \includegraphics[width=0.9\textwidth]{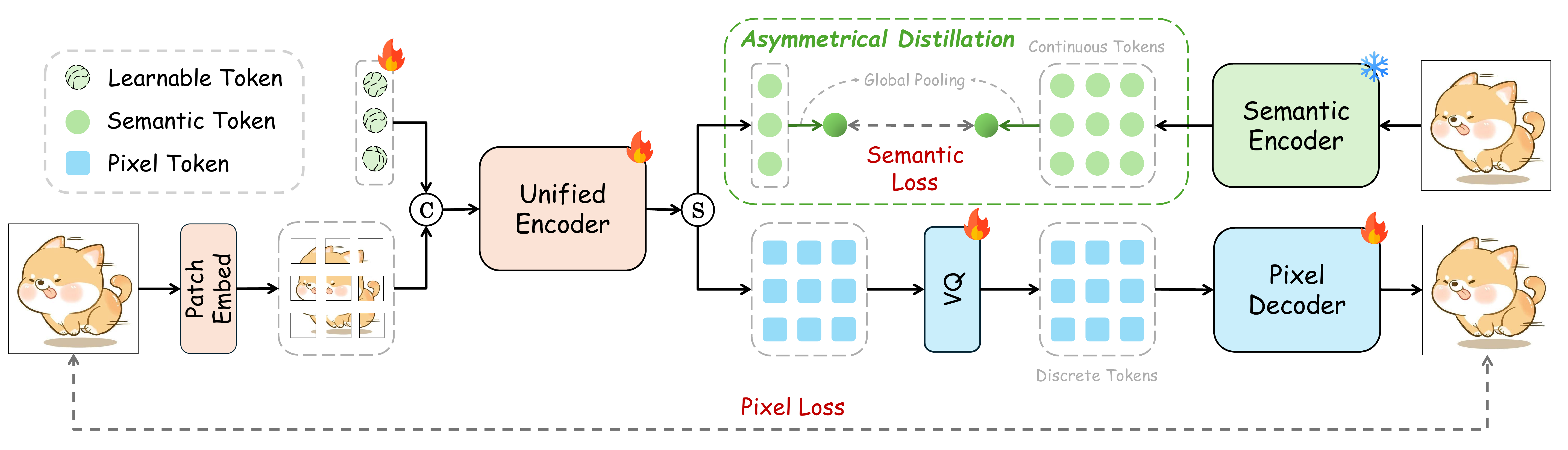}
  \caption{\textbf{Overview of our WinTok.} Our WinTok adopts a hybrid tokenization paradigm integrated with learnable tokens.
  The separate token sets are optimized for visual understanding and generation respectively,
  with asymmetric token distillation for semantic tokens and image reconstruction for pixel tokens.
  Therefore, we mitigate the representation conflict while maintaining a unified tokenizer.
  }
  \label{fig:framework}
\end{figure}

\noindent\textbf{Unified Encoder with Learnable Tokens.}
As depicted in \cref{fig:framework},
WinTok employs a unified encoder \(\mathcal{E}\), structured as a Vision Transformer (ViT) \cite{dosovitskiy2021vit}. 
The input image $\mathbf{X}$ is first patchified into \(N\) non-overlapping patches,
which are subsequently converted into pixel tokens via a patch embedding layer $\mathcal{P}$:
\begin{equation}
\mathbf{P}^{o}=\mathcal{P}(\mathbf{X})=\{\mathbf{P}_1^{o},...,\mathbf{P}_N^{o}\} \label{eq:P}
\end{equation}
However, a singular token set proves inadequate for addressing both visual understanding and generation 
due to the inherent conflict between high-level semantic abstractions and low-level pixel reconstructions. 
To overcome this, we incorporate an additional set of $M$ learnable tokens for task decomposition:
\begin{equation}
\mathbf{S}^{o}=\{\mathbf{S}_1^{o},...,\mathbf{S}_M^{o}\}. \label{eq:S}
\end{equation}
More specifically,
we leverage pixel tokens for visual generation,
since these tokens contain rich local details from the input images.
Alternatively,
we leverage learnable tokens for visual understanding,
since these tokens are the global vectors that can be optimized to summarize the global semantic of the input images.
To make these two types of tokens work collaboratively,
we concatenate them along the sequence dimension and input the combined tokens to our unified encoder:
\(\mathbf{P} \oplus \mathbf{S} = \mathcal{E}(\mathbf{P}^{o} \oplus \mathbf{S}^{o}).\)
This arrangement enables the retention of distinct representations for each token type, 
fostering contextual integration for cooperative functionality. 
The subsequent sections detail the supervision methods for these two token sets.

\setlength{\textfloatsep}{8pt}

\noindent\textbf{Semantic Token: Asymmetric Distillation.} 
Since the enhanced tokens $\mathbf{S}$ derive from the randomly-initialized learnable tokens $\mathbf{S}^{o}$,
we optimize $\mathbf{S}$ to capture high-level semantics for visual understanding.
Specifically,
we propose an asymmetric token distillation technique where we use a visual foundation model $\mathcal{T}$
as the semantic teacher, and extract $K$ semantic tokens of the input image:
\begin{equation}
\mathbf{T}=\mathcal{T}(\mathbf{X})=\{\mathbf{T}_1,...,\mathbf{T}_K\}. \label{eq:T}
\end{equation}
Since the foundation model is well-pretrained with large-scale data,
these tokens $\mathbf{T}$ contain discriminative semantic knowledge to understand the input image.
As such,
we propose to transfer such knowledge from semantic tokens $\mathbf{T}$ to learnable tokens $\mathbf{S}$.
Asymmetry not only lies in the knowledge, but may also exists in the token numbers.
Therefore,
we adopt global pooling on both of them to obtain their corresponding global vectors: 
\(\mathbf{s}=\text{Pool}(\mathbf{S}),~~\mathbf{t}=\text{Pool}(\mathbf{T}).\label{eq:Pool}\)
This process is optimized using a cosine similarity loss:
\begin{equation}
   \mathcal{L}_{sem} = 1 -  \frac{\langle \mathbf{s}, \mathbf{t} \rangle}{||\mathbf{s}||_2 \cdot ||\mathbf{t}||_2}.
  \label{eq:op}
\end{equation}

\noindent\textbf{Pixel Tokens: Image Reconstruction.}
The enhanced tokens $\mathbf{P}$ emerge from the pixel tokens $\mathbf{P}^{o}$ and are utilized for visual generation.
Specifically,
we employ a vector-quantization module \(\mathcal{Q}\) to convert $\mathbf{P}$ into discrete pixel tokens via a learnable codebook:
\begin{equation}
  \mathbf{Q} = \mathcal{Q}(\mathbf{P})=\{\mathbf{Q}_1,...,\mathbf{Q}_N\}. \label{eq:Q}
\end{equation}
These quantized tokens then serve as the input to a decoder \(\mathcal{D}\) to reconstruct the original image:
\begin{equation}
  \hat{\mathbf{X}} = \mathcal{D}(\mathbf{Q}). \label{eq:D}
\end{equation}
We optimize the quantizer and the decoder using a combination of pixel reconstruction and codebook losses \cite{sun2024llamagen,yu2022vqgan}:
\begin{equation}
  \mathcal{L}_{rec} = || \mathbf{X} - \hat{\mathbf{X}} ||_2 + || \text{g}[\mathbf{P}] - \mathbf{Q} ||_2 \\
                                          + \beta || \mathbf{P} - \text{g}[\mathbf{Q}] ||_2,
                                          \label{eq:R}
\end{equation}
where \(\text{g}[\cdot]\) denotes the stop-gradient operation and \(\beta\) is a hyperparameter to balance the loss terms.
Additionally,
we incorporate a perceptual loss \(\mathcal{L}_{per}\) \cite{zhang2018unreasonable} and
an adversarial loss \(\mathcal{L}_{adv}\) \cite{karras2019style} to enhance the visual quality of reconstructed images.
Thus, the optimization of pixel tokens is based on $\mathcal{L}_{pix} = \mathcal{L}_{rec} + \lambda_{per} \mathcal{L}_{per} + \lambda_{adv} \mathcal{L}_{adv}$,
where \(\lambda_{per}\) and \(\lambda_{adv}\) are hyperparameters.

\input{assets/tables/sota_tokenizer.tex}

\noindent\textbf{Training Objectives.}
The overall loss function for optimizing WinTok combines the losses from both token types:
\begin{equation}
  \mathcal{L} =  \mathcal{L}_{sem} + \mathcal{L}_{pix},
\end{equation}
Through asymmetric token distillation, the learnable tokens evolve to encapsulate 
the semantic strengths of the foundation model for visual understanding, 
while pixel tokens learn to capture local details for visual generation. 
Consequently,
our WinTok achieves a win-win performance for both tasks with a hybrid tokenization framework.

\noindent\textbf{WinTok for Downstream Tasks.}
\label{sec:wintok_for_downstream}
WinTok's hybrid structure allows for the effective application of semantic and pixel tokens 
in understanding and generation tasks, respectively.
Therefore, we further integrate WinTok into a pre-trained LLM to excavate its potential in downstream tasks.
For multimodal understanding, the unified encoder \(\mathcal{E}\) extracts continuous semantic tokens \(\mathbf{S}\) from the input image. Then these continuous visual tokens are projected into the textual embedding space with a learnable linear layer, and integrated with text tokens for comprehension.
For visual generation, discrete pixel tokens \(\mathbf{Q}\) are extracted using the unified encoder and quantizer. Subsequently, the LLM learns the joint distribution between these discrete visual tokens given text tokens as condition. We also introduce an autoregressive head to facilitate multi-code prediction as in previous work \cite{lee2022autoregressive, wu2024vila, ma2025unitok}.
During inference, the LLM autoregressively samples discrete visual tokens, which are then decoded into images with the decoder \(\mathcal{D}\).

\input{assets/tables/sota_recon}

%% file: assets/tables/sota_tokenizer.tex
\begin{table}[!t]
\centering
\caption{\textbf{Comparison of reconstruction quality and semantic capability on 256 \(\times\) 256 ImageNet-1K validation set.}
"Capacity" denotes the theoretical combinations of code entries.
* indicates the results are obtained by linear probing.
WinTok achieves state-of-the-art classification accuracy while being competitive in reconstruction,
with significantly less data compared to other unified tokenizers.
}
\resizebox{0.95\textwidth}{!}{%
\begin{tabular}{lccccccc}
\hline
\textbf{Method}                          & \textbf{Type}     & \textbf{Ratio} & \begin{tabular}[c]{@{}c@{}}\textbf{Training}\\ \textbf{Data}\end{tabular} & \begin{tabular}[c]{@{}c@{}}\textbf{Codebook}\\ \textbf{Size}\end{tabular} & \textbf{Capacity} & \textbf{rFID} (\(\downarrow\)) & \textbf{Accuracy} (\(\uparrow\))  \\ \hline
\multicolumn{8}{c}{\textit{\textbf{Semantic Tokenizer}}} \\ \hline
CLIP-L/14 \cite{radford2021clip}         & Continuous        & -              & WIT400M        & - & - & -        & 75.5      \\
Dinov2-L \cite{oquab2023dinov2}          & Continuous        & -              & LVD142M        & - & - & -        & ~~86.3*   \\
SigLIP2-So/16 \cite{tschannen2025siglip} & Continuous        & -              & WebLI10B       & - & - & -        & 83.4      \\ \hline
\multicolumn{8}{c}{\textit{\textbf{Reconstruction-oriented Tokenizer}}}                                                       \\ \hline
SD-VAE 1.x \cite{rombach2022sd}          & Continuous        & 8              & OI1B          & - & - & 1.22        & -       \\
SD-VAE 2.x \cite{rombach2022sd}          & Continuous        & 8              & Mix6B         & - & - & 0.70        & -       \\
SDXL-VAE \cite{podellsdxl}               & Continuous        & 8              & -             & - & - & 0.67        & -       \\
SD-VAE 3.5 \cite{esser2024sd3}           & Continuous        & 8              & -             & - & - & 0.19        & -       \\
FLUX-VAE \cite{flux2024}                 & Continuous        & 8              & -             & - & - & 0.18        & -       \\
VA-VAE \cite{yao2025vavae}               & Continuous        & 16             & IN-1K         & - & - & 0.28        & -       \\
RAE (SigLIP2) \cite{zheng2025rae}        & Continuous        & 16             & IN-1K         & - & - & 0.53        & ~~79.1* \\
VFM-VAE \cite{bi2025vfmvae}              & Continuous        & 16             & IN-1K         & - & - & 0.52        & -       \\
LlamaGen \cite{sun2024llamagen}          & Discrete          & 16             & IN-1K         & 16384 & 2\(^{14}\)  & 2.19  & -       \\
Open-Magvit2 \cite{luo2024open}          & Discrete          & 16             & IN-1K         & 262144 & 2\(^{18}\) & 1.17 & -       \\
VFMTok \cite{zheng2025vfmtok}            & Discrete          & -              & IN-1K         & 16384 & 2\(^{14}\)  & 0.89        & ~~69.4* \\
WeTok \cite{zhuang2025wetok}             & Discrete          & 16             & IN-1K         & - & 2\(^{32}\) & 0.61        & -       \\ 
MGVQ \cite{jia2025mgvq}                  & Discrete          & 16             & IN-1K         & 2048 \(\times\) 8 & 2\(^{88}\) & 0.49        & -       \\ \hline
\multicolumn{8}{c}{\textit{\textbf{Unified Tokenizer}}}  \\ \hline
UniLIP \cite{tang2025unilip}             & Continuous        & 16             & BP-32M        & - & - & 0.79        & - \\
UniFlow (SigLIP2) \cite{yue2025uniflow}  & Continuous        & 16             & IN-1K         & - & - & 0.62        & - \\
VILA-U \cite{wu2024vila}                 & Discrete          & 16             & CY700M         & 16384 & 2\(^{14}\) & 1.80        & 73.3       \\
QLIP-L \cite{zhao2025qlip}               & Discrete          & 16             & DC1B           & - & - & 1.46        & 79.1       \\
DualToken \cite{song2025dualtoken}       & Discrete          & 16             & CC12M          & - & - & 0.54        & 81.6       \\
TokenFlow \cite{qu2025tokenflow}         & Discrete          & 16             & LA+CY700M      & 32768 & 2\(^{15}\) & 1.37        & -          \\
SemHiTok \cite{chen2025semhitok}         & Discrete          & 16             & Mix70M          & 16384 \(\times\) 12 & 2\(^{14}\) & 1.16        & -       \\
TokLIP-L \cite{lin2025toklip}            & Discrete          & 16             & Mix80M         & 16384 & 2\(^{14}\) & 2.19        & 80.0       \\
UniTok \cite{ma2025unitok}               & Discrete          & 16             & DC1B           & 4096 \(\times\) 8   & 2\(^{96}\) & 0.41        & 70.8       \\
VQRAE \cite{du2025vqrae} & Hybrid & 16 & BP-32M & 16384 & 2\(^{14}\) & 1.31 & - \\
\rowcolor[HTML]{EFEFEF} 
{WinTok}                          & {Hybrid}          & 16             & Mix50M          & 4096 \(\times\) 4   & 2\(^{48}\) & {0.41}        & {82.0}       \\ \hline
\end{tabular}%
}
\label{tab:comp_tok}
\end{table}

%% file: assets/tables/sota_recon.tex
\begin{table}[t]
    \centering
    \renewcommand\arraystretch{1.1}
    \caption{\textbf{Comparison of visual reconstruction on ImageNet-1K and MS-COCO 2017 validation set.} Images are resized to 256 \(\times\) 256 for evaluation.
    Our \tok~achieves competitive performance compared to UniTok even with significantly less training data and a reduced codebook size.
    }
    \resizebox{0.8\linewidth}{!}{
        \begin{tabular}{lcccccccc}
        \toprule
        \multirow{2}{*}{\textbf{Method}} & \multirow{2}{*}{\textbf{Ratio}} & \multicolumn{3}{c}{\textbf{MS-COCO 2017}} & & \multicolumn{3}{c}{\textbf{Imagenet-1K}} \\
        \cmidrule{3-5} \cmidrule{7-9}
         & & \textbf{rFID$\downarrow$} & \textbf{PSNR$\uparrow$} & \textbf{SSIM$\uparrow$} & & \textbf{rFID$\downarrow$} & \textbf{PSNR$\uparrow$} & \textbf{SSIM$\uparrow$} \\
        \midrule
        Show-o~\cite{xie2025show} & 16 & 9.26 & 20.90 & 0.59 & & 3.50 & 21.34 & 0.59 \\
        \small{Open-MAGVIT2-I-PT~\cite{luo2024open}} & 16 & 7.93 & 22.21 & 0.62 & & 2.55 & 22.21 & 0.62 \\
        LlamaGen~\cite{sun2024llamagen} & 16 & 8.40 & 20.28 & 0.55 & & 2.47 & 20.65 & 0.54 \\

        WeTok~\cite{zhuang2025wetok} & 16 & 6.55 & 21.99 & 0.63 &  & 1.58 & 22.38 & 0.62 \\
        BSQ~\cite{yang2021bsq} & 16 & - & - & - & & 3.81 & 24.12 & 0.66 \\
        QLIP-B~\cite{zhao2025qlip} & 16 & - & - & - & & 3.21 & 23.16 & 0.63 \\
        TokenFlow~\cite{qu2025tokenflow} & 16 & - & - & - & & 1.37 & 21.41 & {0.69} \\
        UniTok~\cite{ma2025unitok} & 16 & 4.05 & 23.92 & 0.73 &  & 0.41 & 24.31 & 0.71 \\
        \rowcolor[HTML]{EFEFEF} {WinTok} & {16} & {4.24} & {23.83} & {0.72} &  & {0.41} & 24.23 & 0.71 \\

    \bottomrule
    \end{tabular}}
    \label{tab:comp_recon}
\end{table}

%% file: sec/4_exp.tex
\section{Experiment}
\label{sec:exp}

\subsection{Implementation Details}



\noindent\textbf{Tokenizer Setup.}
In our experiments, we adopt ViT-based architectures for both the encoder and the decoder.
The encoder is initialized with SigLIP2-So400M \cite{tschannen2025siglip},
while the decoder is trained from scratch.
The number of learnable tokens is set to 256 by default.
Since the quantization operation is non-differentiable,
we employ the straight-through estimator \cite{bengio2013estimating} to for proper gradient backpropagation.
We adopt Multi-codebook Quantization (MCQ) \cite{ma2025unitok} and 
select SigLIP2-So400M \cite{tschannen2025siglip} as the default semantic teacher.
We use 50M images randomly sampled from open-source datasets \cite{deng2009imagenet, gadre2023datacomp, kakaobrain2022coyo-700m, sharma2018conceptual, changpinyo2021conceptual, wang2025textatlas5m, wang2025faceid}
to train our \tok.
The tokenizer is trained for 5 epochs with global batch size of 256 and learning rate of 2e-4 with warm-up and cosine decay schedule.
All experiments are conducted on H20 GPUs with PyTorch. More details are provided in the supplementary material.

\noindent\textbf{UMM Setup.}
We initialize our unified multimodal model with Qwen3-8B \cite{yang2025qwen3}.
For pre-training stage, we incorporate approximately 80M image-text pairs from \cite{chen2025blip3o, gadre2023datacomp, singla2024pixelprose, schuhmann2022laion}. We further finetune the model using 6M instruction-following data from \cite{liu2024improved, li2024llava} for multimodal understanding, along with
4M synthetic data generated by FLUX.2-klein \cite{flux-2-2025} and Z-Image-Turbo \cite{cai2025zimage} for visual generation.

\noindent\textbf{Evaluation Metrics.}
We evaluate \tok~on ImageNet-1K \cite{deng2009imagenet} validation set
using rFID and Top-1 classification accuracy for state-of-the-art comparison.
For visual reconstruction, we further report rFID, PSNR, and SSIM on ImageNet-1K and MS-COCO 2017 \cite{lin2014microsoft} validation set.
For multimodal understanding, we evaluate on a wide range of benchmarks,
including POPE \cite{li2023pope}, GQA \cite{hudson2019gqa}, TextVQA \cite{singh2019textvqa},
MME-P \cite{fu2024mme}, MMBench \cite{liu2024mmbench}, and MM-Vet \cite{yu2024mmvet}.
For visual generation, we evaluate on GenEval \cite{ghosh2023geneval} and DPG-Bench \cite{hu2024ella}.

\input{assets/tables/sota_mmu.tex}

\subsection{Comparison with State-of-The-Art Methods}

\noindent\textbf{Tokenizer.}
As summarized in \cref{tab:comp_tok}, our \tok~demonstrates superior performance in both reconstruction quality 
and classification accuracy compared to leading visual tokenizers.
In terms of reconstruction quality,
our \tok~surpasses all previous discrete reconstruction-oriented tokenizers as well as most unified tokenizers.
Notably, \tok~achieves a comparable rFID to the state-of-the-art UniTok \cite{ma2025unitok}, 
employing considerably less training data and a reduced codebook size. Moreover, \tok~achieves 100\% codebook usage even with a high capacity.
Regarding semantic representation capabilities,
\tok~outperforms all prior unified tokenizers, exceeding the specialized semantic tokenizer
CLIP-L/14 \cite{radford2021clip} by 6.5\%.
These findings highlight the efficacy of our hybrid tokenization design, 
which incorporates transferable tokens and asymmetric token distillation 
to effectively decompose understanding and generation tasks.

\noindent\textbf{Visual Reconstruction.}
As shown in \cref{tab:comp_recon}, our \tok~achieves impressive reconstruction quality 
on 256 \(\times\) 256 ImageNet-1K and MS-COCO 2017 validation set. 
Notably, \tok~is competitive with the state-of-the-art unified tokenizer UniTok 
while trained on significantly less data (50M vs. 1B) and a reduced codebook capacity (\(2^{48}\) vs. \(2^{96}\)).

\noindent\textbf{Multimodal Understanding.}
As presented in \cref{tab:comp_und}, when integrated with an LLM, \tok~achieves competitive performance
across various multimodal understanding benchmarks.
Our model outperforms several pioneering UMMs,
including SEED-X \cite{ge2024seed} and Liquid \cite{wu2024liquid} on POPE and GQA.
Moreover, we achieve 86.5\% on POPE and 55.2 \% on TextVQA,
surpassing UniTok \cite{ma2025unitok} by 3.3\% and 3.6\%, respectively.
Even with only 64 semantic tokens representing the image, \tok~can still obtain satisfactory comprehension results, outperforming VILA-U \cite{wu2024vila} on MME-P and SemHiTok \cite{chen2025semhitok} on POPE.
Overall, \tok~shows consistent improvements across all benchmarks, indicating its potential for multimodal understanding tasks.

\input{assets/tables/sota_t2i}

\noindent\textbf{Visual Generation.}
As summarized in \cref{tab:comp_t2i}, our multimodal model consistently achieves competitive or even superior performance compared to state-of-the-art diffusion and autoregressive-based models.
Notably, our unified multimodal model outperforms representative autoregressive systems such as Chameleon \cite{team2024chameleon}, LlamaGen \cite{sun2024llamagen}, and Janus \cite{wu2025janus}, while remaining competitive with large-scale diffusion experts trained on billions of image–text pairs.
Furthermore, compared with recent approaches that incorporate unified tokenizers into large language models, \ie, UniTok \cite{ma2025unitok} and TokenFlow \cite{qu2025tokenflow}, WinTok consistently achieves better performance across both benchmarks.
These results highlight the robustness and effectiveness of WinTok as the visual tokenizer within a unified multimodal framework, particularly for complex and compositional text-to-image generation tasks. Moreover, the above observations further confirm that hybrid tokenization substantially benefits downstream unified modeling.

\begin{figure}[!t]
    \centering
    \includegraphics[width=0.95\textwidth]{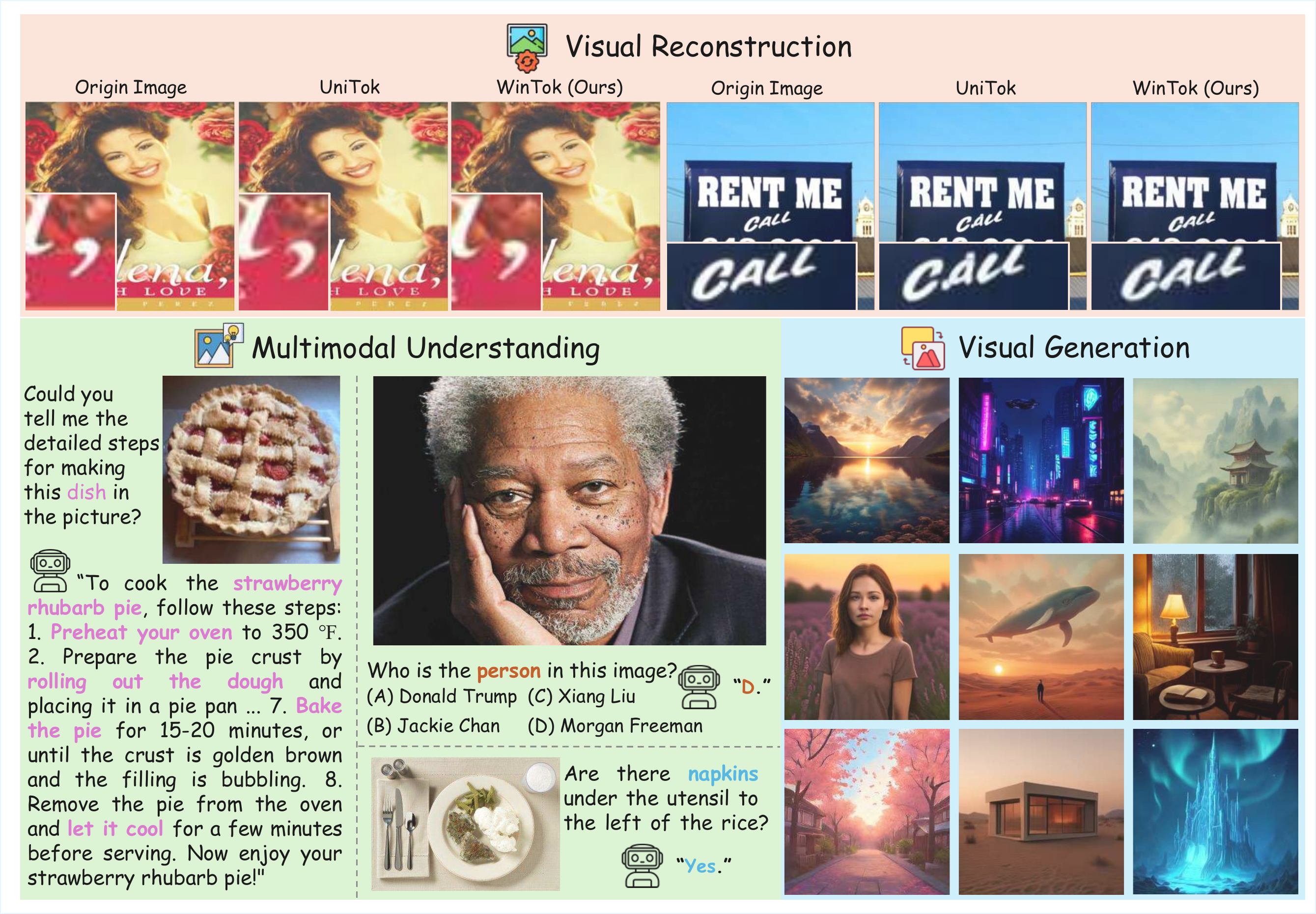}
    \caption{\textbf{Qualitative results demonstrating the superior performance of our \tok~on downstream applications.}}
    \label{fig:vis}
\end{figure}

\noindent\textbf{Visualization} \cref{fig:vis} presents qualitative results of \tok~on visual reconstruction,
multimodal understanding, and visual generation.
Our \tok~not only preserves fine-grained details for high-quality reconstruction,
but also effectively captures global semantic information for accurate multimodal understanding.
Moreover, it enables the generative model to produce diverse and realistic images.
These results substantiate the efficacy of our approach in balancing the 
requirements of both understanding and generation tasks via decomposition with transferable tokens.

\subsection{Comparison with Other Tokenization Strategies}

In this section, we compare our hybrid tokenization approach, WinTok, with several representative visual tokenization strategies for unified visual understanding and generation. Specifically, we consider the following baselines: (a) \textbf{VQVAE} \cite{van2017vqvae}: a pixel-level discrete representation widely used in autoregressive image generation. We adopt the pretrained VQVAE tokenizer from LlamaGen \cite{sun2024llamagen}, which converts images into discrete visual tokens for multimodal modeling. (b) \textbf{Decoupled tokenization}: a dual-encoder design that employs separate visual representations for understanding and generation, using a semantic encoder (SigLIP2 \cite{tschannen2025siglip}) for perception and a VQVAE tokenizer for image synthesis. This strategy resembles previous methods \cite{wu2025janus, deng2025bagel} but has a slightly different implementation. (c) \textbf{Unified tokenizer}: this line of work employs a unified representation that jointly models semantic and pixel-level information. We select UniTok \cite{ma2025unitok} as a representative.

To ensure a fair comparison, we train unified multimodal models with these tokenizers under the same training configuration and data scale. Specifically, we construct a controlled training subset consisting of 10M text-to-image and image-to-text data. All tokenizers are integrated with Qwen3-4B \cite{yang2025qwen3}. All models are evaluated on both visual understanding benchmarks \cite{singh2019textvqa, hudson2019gqa, li2023pope, liu2024mmbench} and text-to-image generation tasks \cite{ghosh2023geneval} to comprehensively assess their performance in unified multimodal modeling.

\begin{figure}[!t]
  \centering
  \begin{subfigure}{0.46\linewidth}
    \centering
    \includegraphics[width=\linewidth]{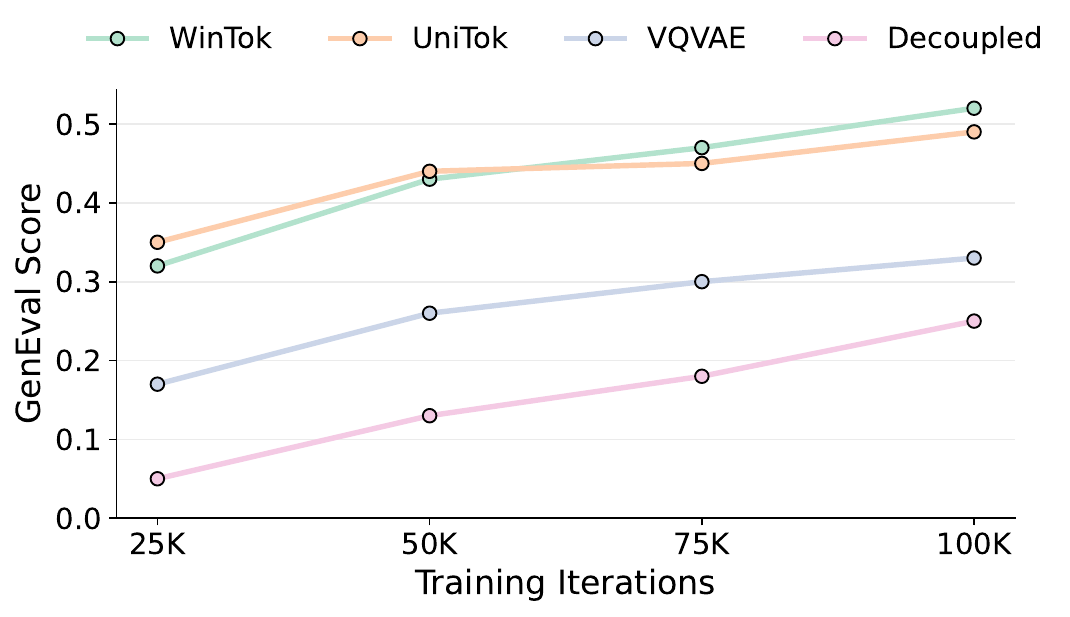}
    \caption{Text-to-image generation performance.}
    \label{fig:comp_tok_gen}
  \end{subfigure}
  \hfill
  \begin{subfigure}{0.52\linewidth}
    \centering
    \includegraphics[width=\linewidth]{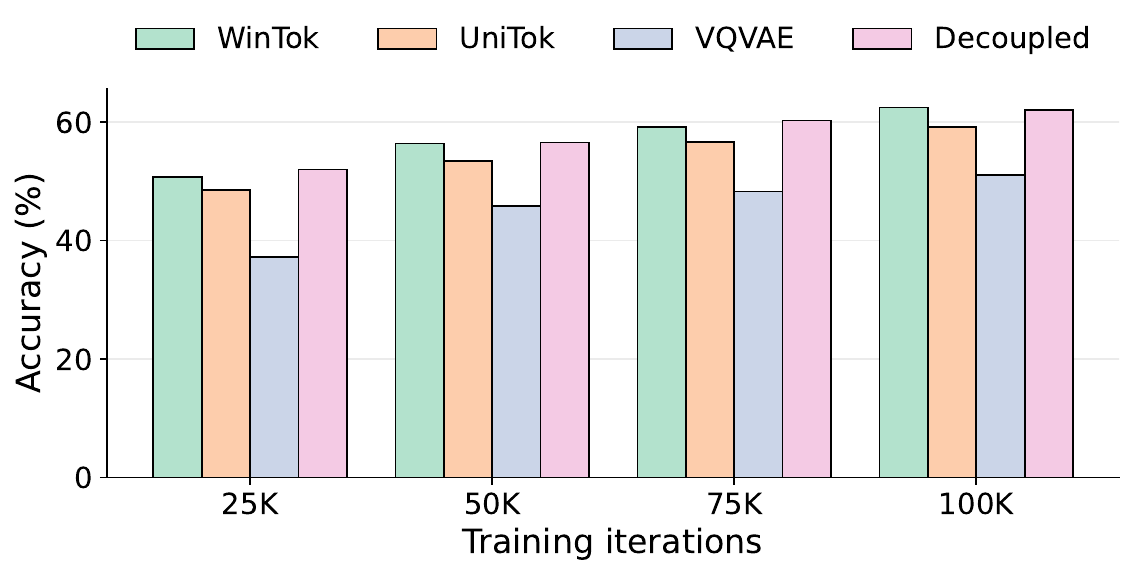}
    \caption{Multimodal understanding performance.}
    \label{fig:comp_tok_und}
  \end{subfigure}
  \caption{\textbf{Comparisons of using different tokenization strategies.} (a) Generation performance is evaluated on GenEval \cite{ghosh2023geneval}. (b) Understanding performance is averaged across 4 benchmarks \cite{singh2019textvqa, hudson2019gqa, li2023pope, liu2024mmbench}.}
  \label{fig:comp_tok_strategy}
\end{figure}

As illustrated in \cref{fig:comp_tok_strategy}, our approach demonstrates strong performance on downstream text-to-image (T2I) generation and multimodal understanding tasks. In T2I generation, our method converges faster and achieves the best results. Compared to the single-stream unified tokenizer UniTok, our method obtains a greater upper bound. In contrast, the VQVAE-based model, which relies solely on pixel-level representation, underperforms in generation. Moreover, the decoupled strategy performs worst in generation, likely due to the inconsistency in representation space and need large-scale training to achieve better performance.

For multimodal understanding, both WinTok and the decoupled strategy achieve satisfactory results owing to the continuous semantic representation. The discrete unified tokenizer UniTok, despite jointly optimizing pixel reconstruction and semantic alignment losses, performs similarly to VQVAE, suggesting that its representation remains biased toward pixel-level information and fails to fully exploit semantic cues. Overall, our method provides the best trade-off between visual generation and understanding, demonstrating strong and balanced performance across both tasks.

\setlength{\textfloatsep}{8pt}

\subsection{Ablations}

\begin{figure}[t]
    \centering
    \begin{subfigure}{0.29\linewidth}
		\centering
		\includegraphics[width=\linewidth]{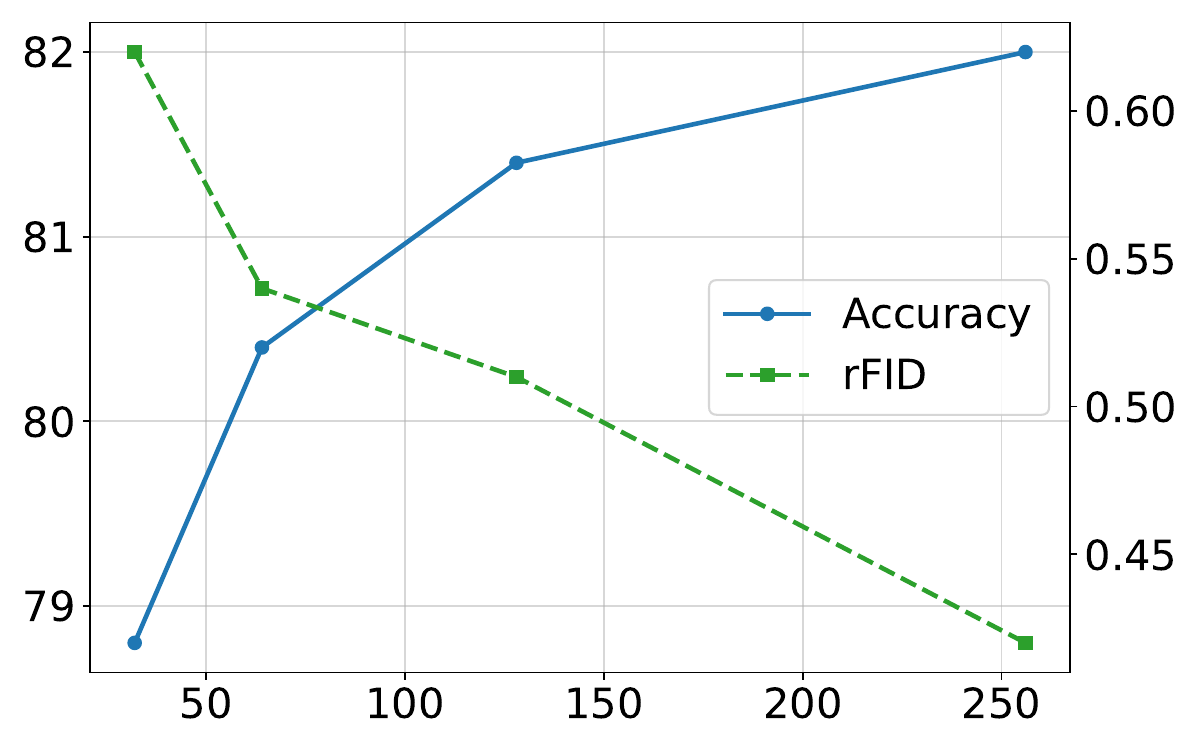}
		\caption{Effects of learnable token numbers.}
		\label{fig:ablation_token_num}
	\end{subfigure}
	\centering
	\begin{subfigure}{0.45\linewidth}
		\centering
		\includegraphics[width=\linewidth]{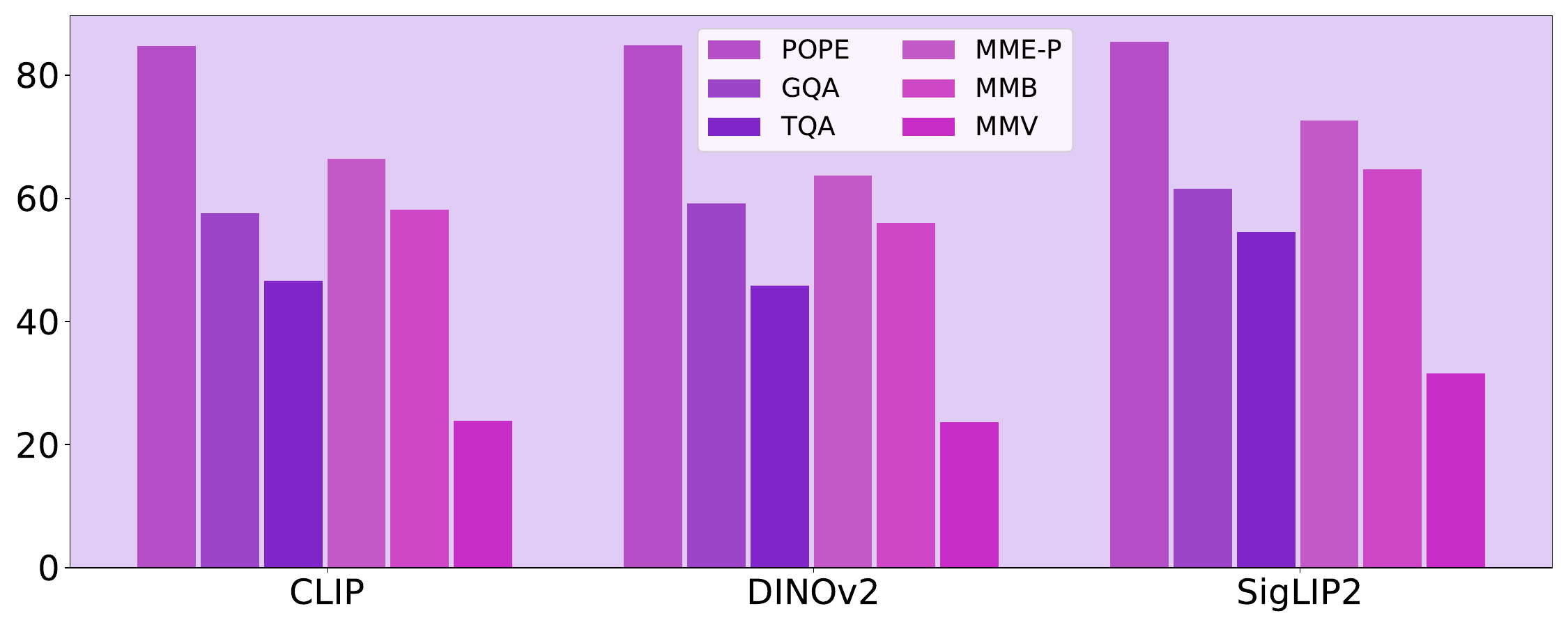}
		\caption{Comparison on using different semantic teachers.}
		\label{fig:ablation_teacher}
	\end{subfigure}
	\centering
	\begin{subfigure}{0.24\linewidth}
		\centering
		\includegraphics[width=\linewidth]{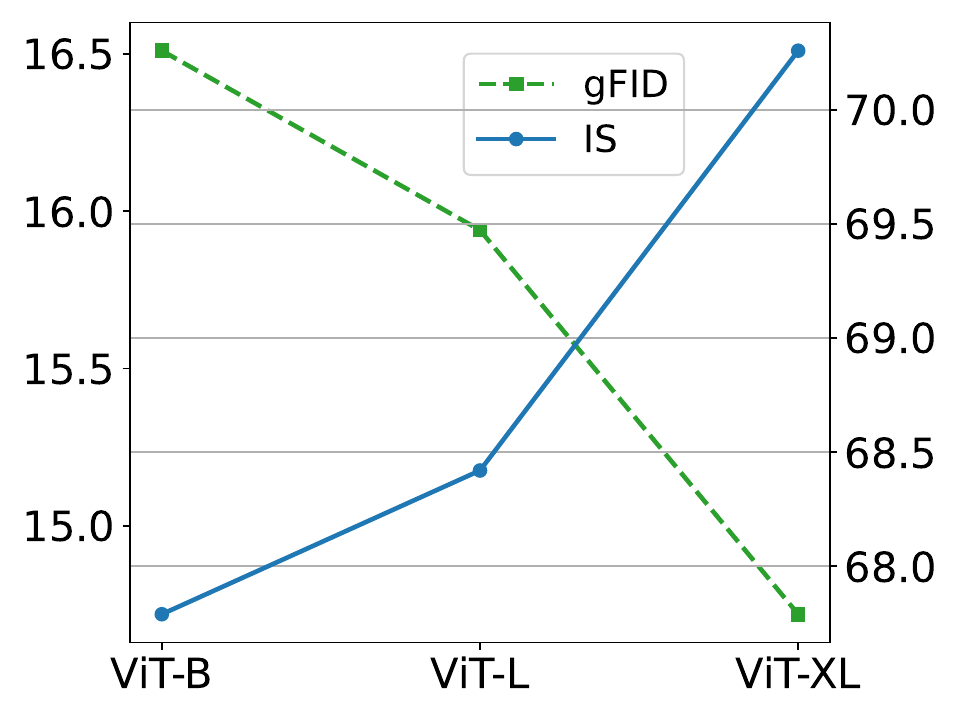}
		\caption{Effects of different decoder sizes.}
		\label{fig:ablation_decoder}
	\end{subfigure}
	\caption{
        \textbf{Ablations on design choices.}(a) As the learnable token number increases, both the reconstruction quality and semantic capability improve.
        (b) Our \tok~demonstrates similar trends when adopting different semantic teachers, while SigLIP2 \cite{tschannen2025siglip} benefits the most.
        (c) A larger decoder not only enhances reconstruction quality but also improves generation performance. Implementation details and 
        more comprehensive analyses are provided in the supplementary material.
        }
	\label{fig:ablation}
\end{figure}

\noindent\textbf{Number of Learnable Tokens.}
We investigate the impact of learnable token quantity on task performance.
As shown in \cref{fig:ablation_token_num},
increasing the number of learnable tokens consistently improves both reconstruction quality and semantic capability.
This improvement arises from the enhanced ability of additional tokens to capture semantic information 
and facilitate the disentanglement of representations, thereby mitigating conflicts between the two tasks.

\noindent\textbf{Semantic Teacher.}
To optimize the transfer of semantic knowledge to the randomly initialized semantic tokens, 
we assess the influence of the chosen semantic teacher model.
We examine three representative visual foundation models: 
CLIP-L/14 \cite{radford2021clip}, DINOv2-L \cite{oquab2023dinov2}, and SigLIP2-So400M \cite{tschannen2025siglip}.
Our results indicate consistent improvements in downstream multimodal understanding tasks across different semantic teachers, 
with SigLIP2 yielding the best performance due to its superior representation capabilities derived from meticulous pre-training.



\noindent\textbf{Decoder Size.}
We explore the effects of varying decoder sizes on both reconstruction quality and generation performance 
by training several \tok~variants.
All variants achieved reasonable reconstruction quality, 
with ViT-B attaining an rFID of 0.60, ViT-L achieving 0.55, and the best-performing ViT-XL achieving 0.54.
As illustrated in \cref{fig:ablation_decoder},
larger decoders not only improve reconstruction quality but also enhance generation performance, 
aligning with findings in prior studies \cite{zheng2025rae, xiong2025gigatok, bachmann2025flextok}.



\subsection{Discussions}

\noindent\textbf{What if we design in the other way?}
To validate our hybrid strategy, we evaluate a reversed variant termed \textit{LoseTok}, where learnable tokens handle reconstruction and pooled pixel tokens manage asymmetric token distillation. As shown in \cref{fig:ablation}, this swap severely degrades reconstruction quality while barely affecting semantic capability. This drop in generative performance occurs because learnable tokens lack the capacity to preserve fine-grained spatial details. While this finding appears to contrast with some prior works \cite{bachmann2025flextok, li2024imagefolder}, the discrepancy arises from the additional semantic supervision imposed on pixel tokens in our setting. Ultimately, this ablation confirms the \tok~design, highlighting the necessity of strictly disentangling semantic abstraction from pixel-level reconstruction.

\begin{figure}[!t]
    \centering
    \begin{subfigure}{0.5\linewidth}
		\centering
		\includegraphics[width=\linewidth]{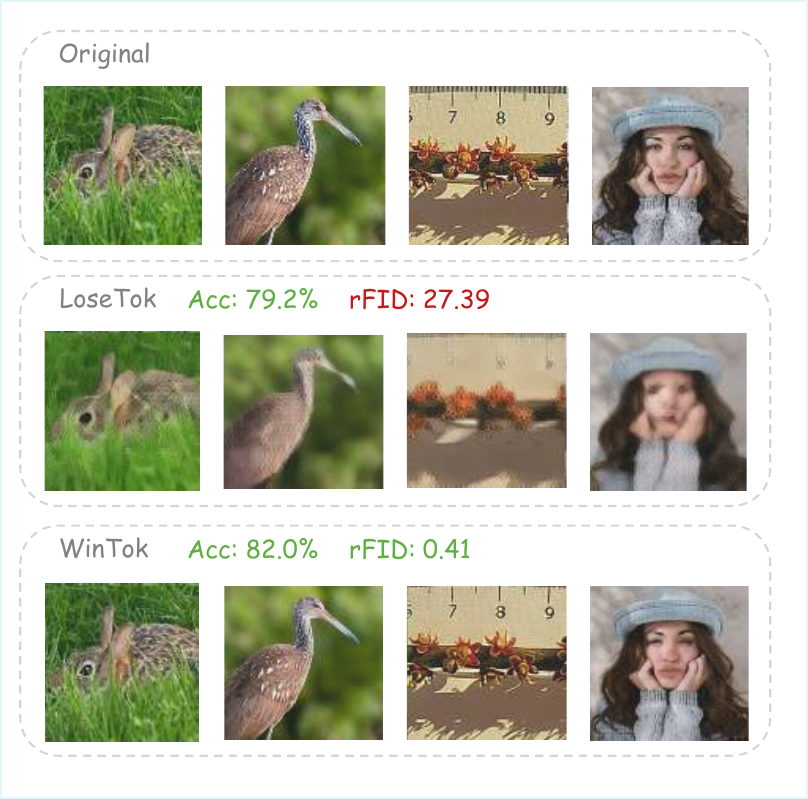}
		\caption{Alternative design consideration.}
		\label{fig:ablation}
	\end{subfigure}
	\centering
	\begin{subfigure}{0.44\linewidth}
		\centering
		\includegraphics[width=\linewidth]{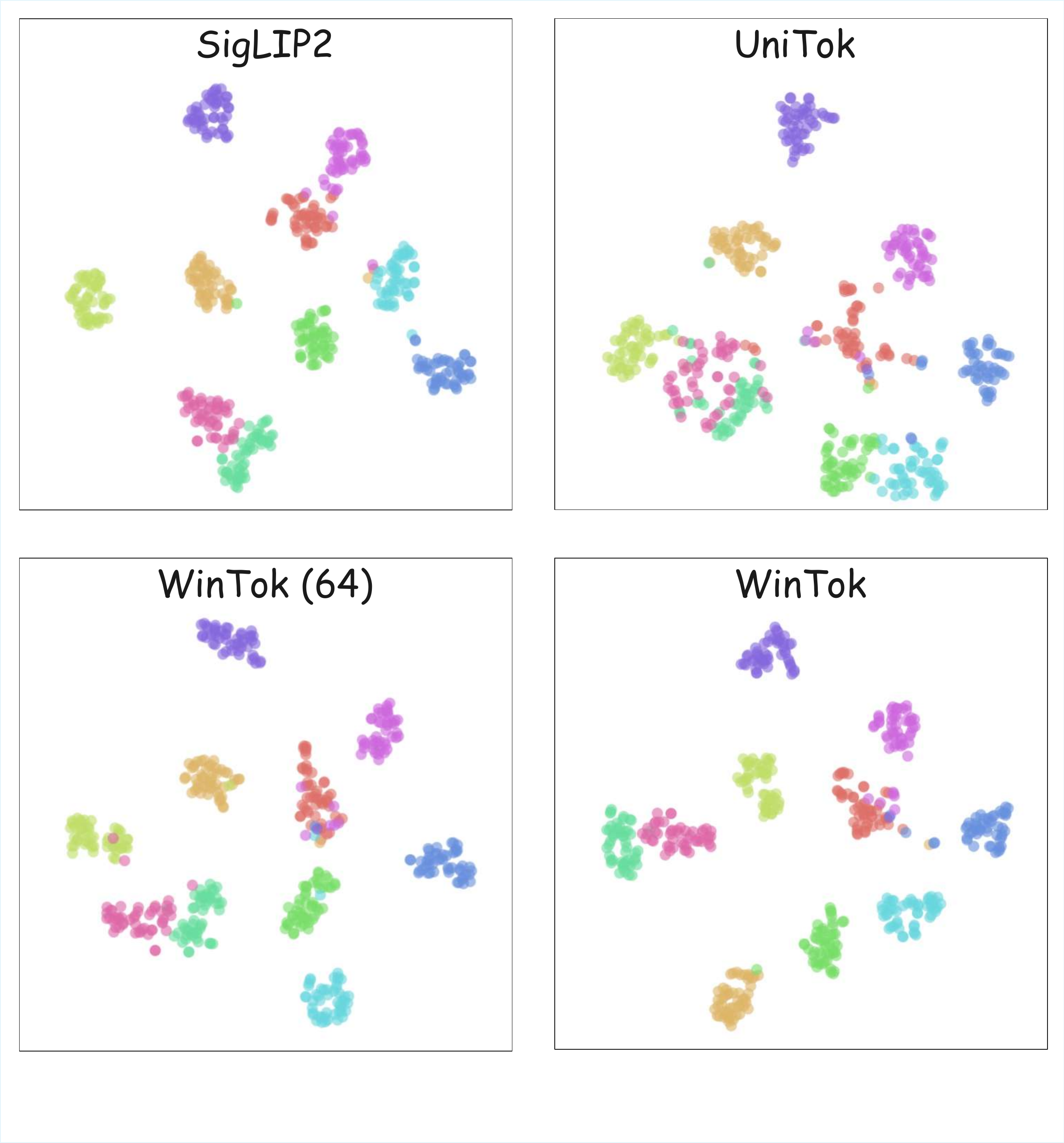}
		\caption{Clusters of different tokenizers.}
		\label{fig:tsne}
	\end{subfigure}
	\caption{
        \textbf{Discussions.} (a) Reversing the roles of the two token types results in significant performance degradation, and WinTok is more rationale.
        (b) WinTok produces more discriminative clusters compared to other tokenizers, even with only 64 tokens representing the global semantics.
        }
	\label{fig:discussion}
\end{figure}

\noindent\textbf{Do the transferable tokens truly learn?}
To evaluate the effectiveness of learnable tokens in capturing transferable semantic representations, we visualize the t-SNE \cite{van2008tsne} embeddings of the learned semantic tokens alongside the pooled image features produced by pre-trained tokenizers, including SigLIP2 \cite{tschannen2025siglip} and UniTok \cite{ma2025unitok}.
As shown in \cref{fig:tsne}, the semantic tokens generated by \tok~form more compact and well-separated clusters than those obtained from the other two tokenizers, despite using only 64 tokens to encode global semantics. This result highlights the superiority of our asymmetric token distillation strategy in effectively transferring semantic knowledge.

\setlength{\textfloatsep}{8pt}

%% file: assets/tables/sota_mmu.tex
\begin{table*}[!t]
\centering
\caption{\textbf{Evaluation on multimodal understanding benchmarks.} \tok~achieves superior performance compared to other unified tokenizers integrated with LLMs
or UMMs. For instance, we outperform TokenFlow-L by 14.6\% on MMBench and UniTok by 3.3\% on POPE.
{WinTok\(^\dagger\)} denotes using only {64} semantic tokens.
}
\resizebox{\textwidth}{!}{%
\begin{tabular}{lcccccccccc}
\hline
\textbf{Method} & \textbf{LLM} & \textbf{Token Type} & \textbf{Res.} & \textbf{POPE} & \textbf{GQA} & \textbf{TextVQA} & \textbf{MME-P} & \textbf{MMBench} & \textbf{MM-Vet} \\ \hline
\multicolumn{10}{c}{\textit{\textbf{Understanding Only MLLM}}}                     \\ \hline
InstructBLIP \cite{dai2023instructblip}  & Vicuna-7B  & 2D-Continuous    & 224 & -    & 49.2 & 50.7 & -      & -    & 26.2 \\
ShareGPT4V \cite{chen2024sharegpt4v}     & Vicuna-7B  & 2D-Continuous    & 336 & -    & 63.3 & 60.4 & 1567.4 & 68.8 & 37.6 \\
LLaVA-v1.5 \cite{liu2024improved}        & Vicuna-7B & 2D-Continuous    & 336 & 85.9 & 62.0 & 58.2 & 1510.7 & 64.3 & 30.5 \\
Qwen2.5-VL \cite{bai2025qwen2}           & Qwen2.5-7B & 2D-Continuous    & dynamic & - & - & 84.9 & - & 83.5 & 67.1 \\
InternVL2.5 \cite{chen2024expanding}     & InternLM2.5-7B  & 2D-Continuous    & dynamic & 90.6 & - & 79.1 & - & 84.6 & 62.8 \\
LLaVA-OneVision \cite{li2024llava}       & Qwen2-7B  & 2D-Continuous    & 384 & - & - & - & 1580.0 & 80.8 & 57.5 \\ \hline
\multicolumn{10}{c}{\textit{\textbf{Unified Multimodal Model}}}                    \\ \hline
LWM \cite{liu2024world}                  & LLaMA2-7B & 2D-Discrete      & 256 & 75.2 & 44.8 & 18.8 & -      & -    & 9.6  \\ 
Show-o \cite{xie2025show}                & Phi-1.5-1.3B  & 2D-Discrete      & 256 & 80.0 & 58.0 & -    & 1097.2 & -    & -    \\ 
Liquid \cite{wu2024liquid}               & Gemma-7B & 2D-Discrete      & 512 & 81.1 & 58.4 & 42.4 & 1119.0 & -    & -    \\ 
Emu3 \cite{wang2024emu3}                 & 8B (from scratch) & 2D-Discrete & 512 & 85.2 & 60.3 & 64.7 & 1243.8 & 58.5 & 37.2 \\
Janus-Pro \cite{chen2025januspro}        & DeepSeek-LLM-7B & 2D-Continuous & 384 & 87.4 & 62.0 & -  & 1567.1 & 79.2 & 50.0 \\ 
LaVIT \cite{jin2024unified}              & LLaMA-7B & 2D-Continuous    & 224 & -    & 46.8 & -    & -      & 58.0 & -    \\ 
SEED-X \cite{ge2024seed}                 & LLaMA2-13B & 2D-Continuous    & 448 & 84.1 & 49.1 & -    & 1457.0 & 70.1 & 43.0 \\
ILLUME \cite{wang2025illume}             & Vicuna-7B & 2D-Continuous    & 224 & 88.5 & -    & 72.1 & 1445.3 & 65.1 & 37.0 \\
VARGPT \cite{zhuang2025vargpt}           & Vicuna-7B & 2D-Continuous  & 256 & 85.9 & -    & -    & 1488.8 & 67.6 & -    \\
BLIP3-o \cite{chen2025blip3o}            & Qwen2.5VL-7B-Instruct & 2D-Continuous    & dynamic     & - & - & 83.1 & 1682.6 & 83.5 & 66.6 \\ \hline
\multicolumn{10}{c}{\textit{\textbf{Unified Tokenizer w/ LLM}}}                  \\ \hline
VILA-U \cite{wu2024vila}                 & LLaMA2-7B & 2D-Discrete     & 256 & 83.9 & 58.3 & 48.3 & 1336.2 & -    & 27.7 \\ 
UniTok \cite{ma2025unitok}               & LLaMA2-7B & 2D-Discrete     & 256 & 83.2 & 61.1 & 51.6 & 1448.0 & -    & 33.9 \\
MUSE-VL \cite{xie2025muse}               & Qwen2.5-7B & 2D-Discrete & 256 & - & - & - & - & 72.1 & - \\
TokLIP \cite{lin2025toklip}              & Qwen2.5-7B-Instruct & 1D-Continuous & 384 & 84.9 & 57.0 & - & 1496.6 & 76.9 & - \\
TokenFlow-L \cite{qu2025tokenflow}       & Vicuna-13B & 2D-Discrete      & 256 & 85.0 & 60.3 & 54.1 & 1365.4 & 60.3 & 27.7 \\
SemHiTok \cite{chen2025semhitok}         & Qwen2.5-7B-Instruct & 2D-Discrete     & 256 & 83.4 & 60.3 & -    & 1449.0 & 72.3    & 30.5    \\
VQRAE \cite{du2025vqrae} & Vicuna-13B & 2D-Continuous & 256 &  85.1 & 63.4 & 46.5 & 1491.1 & 65.5 & - \\
\rowcolor[HTML]{EFEFEF} 
{WinTok\(^{\dagger}\)}                   & Qwen3-8B & 1D-Continuous   & 256 & 84.1 & 58.5 & 47.1 & 1370.4 & 60.2 & 25.2 \\
\rowcolor[HTML]{EFEFEF} 
{WinTok}                                 & Qwen3-8B & 1D-Continuous   & 256 & {86.5} & {62.4} & {55.2} & {1552.0} & {74.9} & {34.6} \\ \hline
\end{tabular}%
}
\label{tab:comp_und}
\end{table*}

%% file: assets/tables/sota_t2i.tex
\begin{table}[t]
    \centering
    \renewcommand\arraystretch{1.1}
    \caption{\textbf{Comparison of visual generation on GenEval and DPG-Bench.}
    }
    \resizebox{0.9\linewidth}{!}{
        \begin{tabular}{lcccccccccc}
        \toprule
        \multirow{2}{*}{\textbf{Method}} & \multirow{2}{*}{\textbf{\# Params}} & \multicolumn{4}{c}{\textbf{GenEval}} & \multicolumn{4}{c}{\textbf{DPG-Bench}} \\
        \cmidrule{3-6} \cmidrule(lr){7-10} 
        & & \textbf{Two Obj.} & \textbf{Counting} & \textbf{Color Attri.} & \textbf{Overall$\uparrow$} 
        & \textbf{Entity} & \textbf{Attribute} & \textbf{Relation} & \textbf{Overall$\uparrow$} \\
        \midrule
        \multicolumn{10}{c}{\textit{\textbf{Diffusion-based Model}}} \\
        \midrule
        SDv1.5~\cite{rombach2022sd} & 0.9B & 0.38 & 0.35 & 0.06 & 0.43 & 74.23 & 75.39 & 73.49 & 63.18 \\
        SDXL~\cite{podellsdxl} & 2.6B & 0.74 & 0.39 & 0.23 & 0.55 & 82.43 & 80.91 & 86.76 & 74.65 \\
        PixArt-$\alpha$~\cite{chen2023pixart} & 0.6B & 0.50 & 0.44 & 0.07 & 0.48 & 79.32 & 78.60 & 82.57 & 71.11 \\
        PixArt-$\Sigma$~\cite{chen2024pixart} & 0.6B & - & - & - & - & 82.89 & 88.94 & 86.59 & 80.54 \\
        Hunyuan DiT~\cite{li2024hunyuan} & 1.5B & - & - & - & - & 80.59 & 88.01 & 74.36 & 78.87 \\
        DALLE3~\cite{betker2023dalle3} & - & 0.87 & 0.47 & 0.45 & 0.67 & 89.61 & 88.39 & 90.58 & 83.50 \\
        SD3-Medium~\cite{esser2024sd3} & 2B & 0.94 & 0.72 & 0.60 & 0.74 & 91.01 & 88.83 & 80.70 & 84.08 \\
        SANA-1.5~\cite{xie2025sana} & 4.8B & 0.93 & 0.86 & 0.65 & {0.81} & - & - & - & 84.70 \\
        \midrule
        \multicolumn{10}{c}{\textit{\textbf{Autoregressive-based Model}}} \\
        \midrule
        Chameleon~\cite{team2024chameleon} & 7B & - & - & - & 0.39 & - & - & - & - \\
        LlamaGen~\cite{sun2024llamagen} & 0.8B & 0.34 & 0.21 & 0.04 & 0.32 & 75.43 & 76.17 & 84.76 & 64.84 \\
        Janus~\cite{wu2025janus} & 1.3B & 0.68 & 0.30 & 0.42 & 0.61 & 87.38 & 87.70 & 85.46 & 79.68 \\
        Harmon~\cite{wu2025harmonizing} & 1.5B & 0.86 & 0.66 & 0.48 & 0.76 & - & - & - & - \\
        Transfusion~\cite{zhou2025transfusion} & 7B & - & - & - & 0.63 & - & - & - & - \\
        
        UniTok~\cite{ma2025unitok} & 7B & 0.73 & 0.38 & 0.41 & 0.59 & 88.27 & 88.05 & 88.45 & 81.18 \\
        TokenFlow~\cite{qu2025tokenflow} & 13B & 0.66 & 0.40 & 0.26 & 0.55 & 79.22 & 81.29 & 85.22 & 73.38 \\
        
        \rowcolor[HTML]{EFEFEF} {WinTok} & {8B} & {0.88} & {0.75} & {0.55} & 0.76 & {89.51} & 88.27 & 89.78 & 83.36 \\

    \bottomrule
    \end{tabular}}
    \label{tab:comp_t2i}
\end{table}

%% file: sec/5_conclusion.tex
\section{Conclusion}

This work presents \tok, a hybrid tokenizer that balances visual understanding and generation by using learnable tokens for global semantic distillation and pixel tokens for local detail reconstruction. Through such hybrid encoding, \tok~can provide effective and flexible
representations for downstream unified modeling. While experiments demonstrate its superior performance as a versatile foundation for unified multimodal models, its current generalization is constrained by a modest 50M-sample training dataset and a lack of downstream architectural exploration beyond Qwen3-8B. Future research will focus on scaling the training corpus to billion-scale levels and co-designing novel unified architectures to fully leverage \tok's unique hybrid representations.

%% file: sec/X_suppl.tex
\clearpage
\setcounter{page}{1}

\begin{minipage}[c]{0.5\textwidth}
\centering
\captionof{table}{\textbf{Training settings of WinTok.}\label{tab:setting_tok}}
\resizebox{\columnwidth}{!}{%
\begin{tabular}{lc}
\hline
Initialization & SigLIP2-So400M-Patch16 \\
Training Data & Mix50M \\
Resolution & 256 \(\times\) 256 \\ 
Data Augmentation & Random Crop \& Resize \\ 
Downsample Ratio & 16 \\ 
EMA & False \\
Transformer Depth & 27 \\ 
Transformer Width & 1152 \\
Codebook Size & 4 \(\times\) 4096 \\
Codebook Dimension & 32 \\
Number of Learnable Tokens & 256 \\
Optimizer & AdamW \\
Optimizer Momentum & (0.9, 0.95) \\
Learning Rate Scheduler & Cosine Decay \\
Base Learning Rate & 2e-4 \\
Warnup Epochs & 0.1 \\
Weight Decay & 0.02 \\
Global Batch Size & 256 \\
Total Epochs & 5 \\
GPUs & 32 H20 \\ \hline
\end{tabular}%
}
\end{minipage}
\begin{minipage}[c]{0.5\textwidth}
\centering
\scriptsize
\captionof{table}{\textbf{Training settings of UMM.}\label{tab:setting_mllm}}
\resizebox{\columnwidth}{!}{%
\begin{tabular}{lc}
\hline
Training Stages & 2 Stages \\
Training Data & Pretrain (80M) \& SFT (10M) \\
Vision Tokenizer & WinTok \\
LLM & Qwen3-8B \\
Optimizer & AdamW \\
Optimizer Momentum & (0.9, 0.999) \\
Weight Decay & 0.0 \\
Warnup Ratio & 0.03 \\
Max Length & 2048 \\
Learning Rate Scheduler & Cosine \\
Base Learning Rate & 1e-4 \& 8e-5 \\
Global Batch Size & 256 \\
Total Epochs & 1 \\
GPUs & 256 H20 \\ \hline
\end{tabular}%
}
\end{minipage}

\section{Additional Implementation Details}
\label{supp:impl}
\subsection{Motivation}
As shown in \cref{tab:rep_conflict}, we conduct experiments to verify whether a single visual tokenizer
can simultaneously satisfy the requirements of both visual understanding and generation tasks.
We select two representative tokenizers:
the semantic tokenizer SigLIP2-So400M-Patch16 \cite{tschannen2025siglip}
and the pixel tokenizer WeTok \cite{zhuang2025wetok} that trained on ImageNet-1K.
We adapt SigLIP2 by adding the pixel decoder same as ours and train the model with reconstruction loss.
We also maintain a semantic consistency loss to preserve the semantic information.
For WeTok, we extract its visual features before quantization and align it with SigLIP2 features using a
cosine similarity loss, while keeping the reconstruction loss.

\subsection{Tokenizer}
\noindent\textbf{Training Data.}
The training data listed in Table~\ref{tab:comp_tok} are detailed as follows:
WIT400M \cite{radford2021clip}, LVD142M \cite{oquab2023dinov2}, WebLI10B \cite{chen2023pali},
OI1B (OpenImages) \cite{kuznetsova2020open}, Mix6B (A mixture of LAION-Aesthetics and LAION-Humans) \cite{schuhmann2022laion},
IN-1K (ImageNet-1K) \cite{deng2009imagenet}, BP-32M (BLIP3o-Pretrain-32M) \cite{chen2025blip3o}, CY700M (COYO-700M) \cite{kakaobrain2022coyo-700m},
DC1B (DataComp-1B) \cite{gadre2023datacomp}, CC12M \cite{changpinyo2021conceptual},
LA+CY700M (A Mixture of LAION and COYO-700M), Mix70M \cite{chen2025semhitok} (A Mixture of 50M subset of COYO-700M, ImageNet-1K, and 20M  MidJourney-style
synthetic data), Mix80M \cite{lin2025toklip} (A mixture of 80M samples from CapsFusion, 
CC12M, and LAION-High-Resolution), and Mix50M (A mixture of 50M samples from ImageNet-1K \cite{deng2009imagenet}, DataComp \cite{gadre2023datacomp}, CC3M \cite{sharma2018conceptual}, CC12M \cite{changpinyo2021conceptual}, COYO \cite{kakaobrain2022coyo-700m}, TextAtlas5M \cite{wang2025textatlas5m}, and FaceID-6M \cite{wang2025faceid}).

\noindent\textbf{WinTok Implementation Details.}
We adopt ViT-based encoder-decoder architecture for WinTok.
The encoder is initialized from SigLIP2-So400M \cite{tschannen2025siglip},
and the decoder is trained from scratch. Both the encoder and decoder share the same architecture
with 27 layers, 16 attention heads, and a hidden dimension of 1152.
The quantizer is implemented using Multi-codebook Quantization (MCQ) \cite{ma2025unitok}
with 4 codebooks, each containing 4096 entries.
We set the number of learnable tokens to 256 by default, and add 1D positional embeddings to these tokens.
On the decoder side, we also add 2D sinosoidal positional embeddings to the pixel tokens.
We provide the detailed training settings of WinTok in Table~\ref{tab:setting_tok}.

\noindent\textbf{Ablations}
For all of the ablation experiments, we train WinTok for 20 epochs with a global batch size of 256 on ImageNet-1K.
We only vary the specific components under study while keeping all other settings consistent with the default configuration.
As for Fig.~\ref{fig:ablation_token_num}, we vary the number of learnable tokens with values of \{32, 64, 128, 256\}.
While for Fig.~\ref{fig:ablation_teacher}, we experiment with three different teacher models: 
CLIP-ViT-L/14 \cite{radford2021clip}, Dinov2-L/14 \cite{oquab2023dinov2}, and SigLIP2-So400M-Patch16 \cite{tschannen2025siglip}.
Since CLIP and Dinov2 accept images of resolution 224\(\times\)224, we resize the input images accordingly when computing the teacher features.
For Fig.~\ref{fig:ablation_decoder}, we compare three different sizes of decoders:
ViT-B (12 layers, 12 heads, 768 hidden dim),
ViT-L (24 layers, 16 heads, 1024 hidden dim), and
ViT-XL (27 layers, 16 heads, 1152 hidden dim).



\subsection{Unified Multimodal Model}
As depicted in \cref{sec:wintok_for_downstream}, we adopt WinTok's continuous semantic tokens for multimodal understanding task and discrete pixel tokens for generation task. We integrate WinTok into a pretrained LLM Qwen3-8B and train with the standard next-token prediction loss. The detailed training recipe in shown in \cref{tab:setting_mllm}.


\section{Additional Quantitative Results}

\noindent\textbf{Effects of learnable token number.}
Quantitative results of Fig~\ref{fig:ablation_token_num} as well as downstream understanding performance on MME-P
are provided in Table~\ref{tab:ablation_token_num}.
As the number of learnable tokens increases, both reconstruction quality and downstream understanding performance improve.
This demonstrates that a larger number of learnable tokens can provide more capacity to capture semantic information,
while benefiting the decoupling of semantic and pixel information, thereby mitigating the representation conflict.

\begin{table}[!htbp]
\caption{\textbf{Quantitative results of using different learnable token numbers.}
Default setting is marked in \colorbox{Gray}{gray}.}
\label{tab:ablation_token_num}
{%
\begin{tabular}{cccc}
\hline
\textbf{\# Learnable Tokens} & \textbf{Accuracy} & \textbf{rFID} & \textbf{MME-P}  \\ \hline
32                  & 78.8     & 0.55 & 1248.0 \\
64                  & 80.4     & 0.51 & 1330.9 \\
128                 & 81.4     & 0.52 & 1350.0 \\
\rowcolor[HTML]{EFEFEF}
256                 & 82.0     & 0.42 & 1453.0 \\ \hline
\end{tabular}%
}
\end{table}

\noindent\textbf{Comparison of using different semantic teachers.}
Quantitative results of Fig~\ref{fig:ablation_teacher} are shown in Table~\ref{tab:ablation_teacher}.
Results of MME-P are divied by 20 for better visualization.

\begin{table}[!htbp]
\caption{\textbf{Quantitative results of using different semantic teachers.}
We use \colorbox{Gray}{SigLIP2-So400M} \cite{tschannen2025siglip} as the semantic teacher by default as such setting achieves the best performance
across all downstream understanding benchmarks.
}
\label{tab:ablation_teacher}
{%
\begin{tabular}{lcccccc}
\hline
\textbf{Teacher} & \textbf{POPE} & \textbf{GQA} & \textbf{TQA} & \textbf{MME-P} & \textbf{MMB} & \textbf{MMV} \\ \hline
CLIP-L/14        & 84.7          & 57.6         & 46.6         & 1328.4         & 58.1         & 23.9         \\
Dinov2-L         & 84.9          & 59.2         & 45.8         & 1274.5         & 56.0         & 23.6         \\
\rowcolor[HTML]{EFEFEF} 
SigLIP2-So400M   & 85.4          & 61.5         & 54.5         & 1453.0         & 64.7         & 31.6         \\ \hline
\end{tabular}%
}
\end{table}

\noindent\textbf{Effects of different decoder sizes.}
Quantitative results of Fig~\ref{fig:ablation_decoder} are provided in Table~\ref{tab:ablation_decoder}.
As can be seen, a larger decoder size leads to better reconstruction quality,
which demonstrates the importance of a powerful decoder in our WinTok framework.
Moreover, a larger decoder also benefits the downstream generation performance.

\begin{table}[!htbp]
\caption{\textbf{Quantitative results of using different decoder sizes.}
Default setting is marked in \colorbox{Gray}{gray}.
}
\label{tab:ablation_decoder}
{%
\begin{tabular}{lccccc}
\hline
\textbf{Variant} & \textbf{\# Params. (M)} & \textbf{GFLOPs} & \textbf{rFID (\(\downarrow\))} & \textbf{gFID (\(\downarrow\))} & \textbf{IS (\(\uparrow\))} \\ \hline
ViT-B            & 92                      & 237             & 0.60          & 16.51         & 67.79       \\
ViT-L            & 315                     & 293             & 0.55          & 15.94         & 68.42       \\
\rowcolor[HTML]{EFEFEF} 
ViT-XL           & 427                     & 321             & 0.54          & 14.72         & 70.26       \\ \hline
\end{tabular}%
}
\end{table}

\section{Additional Qualitative Results}
\subsection{Visual Reconstruction}
Fig.~\ref{fig:vis_2} presents more qualitative results of visual reconstruction from different visual tokenizers.
Our WinTok effectively preserves both semantic and pixel-level details,
especially on textual and facial regions, demonstrating its effectiveness in achieving
high-fidelity visual reconstruction, despite using a relatively small training dataset.

\subsection{Multimodal Understanding}
\begin{figure}
    \centering
    \includegraphics[width=0.8\textwidth]{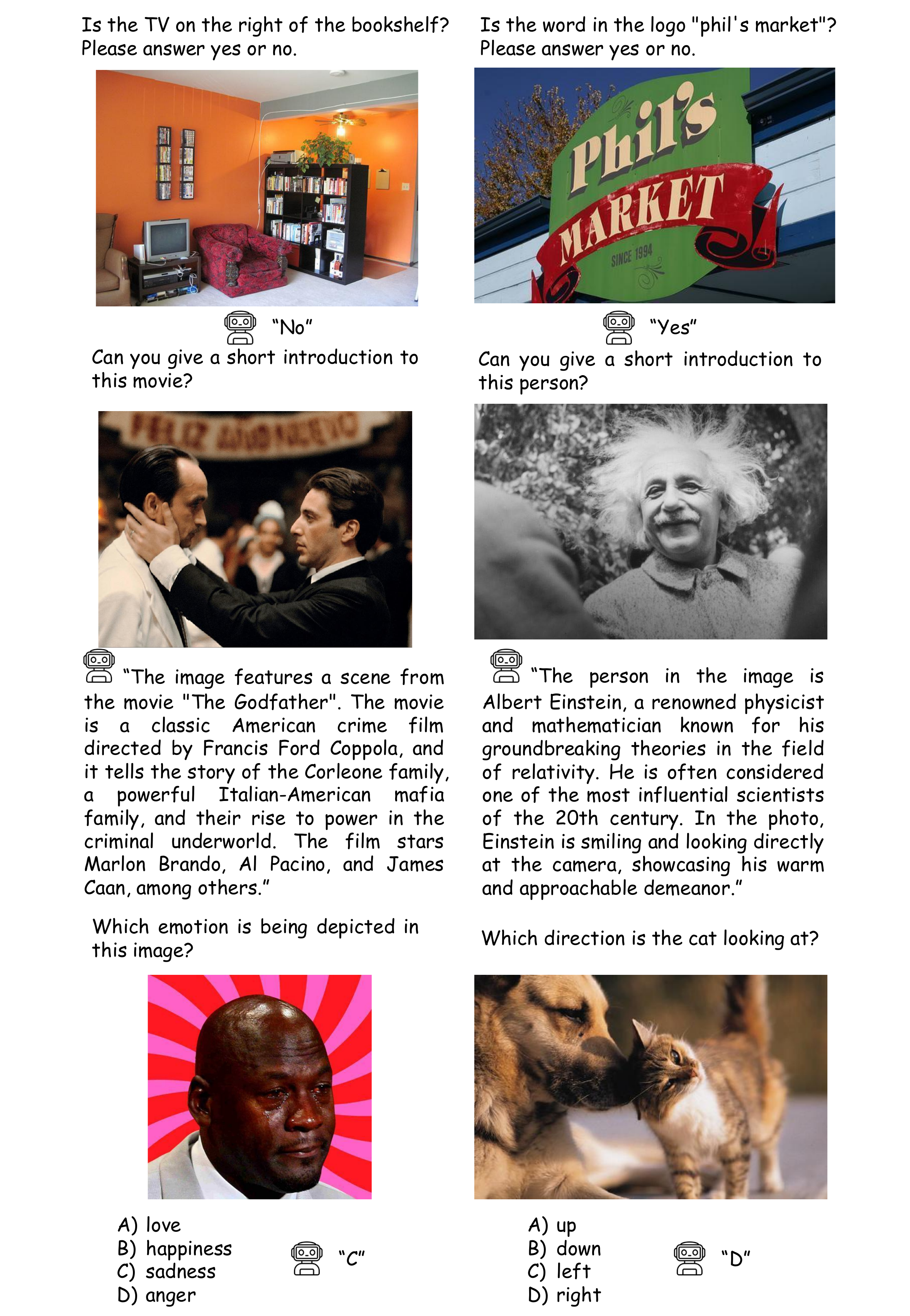}
    \caption{\textbf{Qualitative results of multimodal understanding.}}
    \label{fig:vis_und}
\end{figure}

We provide more qualitative results of multimodal understanding in Fig.~\ref{fig:vis_und}.
Our WinTok-based MLLM can accurately comprehend and answer various types of questions,
including spatial relationships, complex reasoning, and fine-grained attribute recognition.

\subsection{Visual Generation}
\begin{figure}
    \centering
    \includegraphics[width=\textwidth]{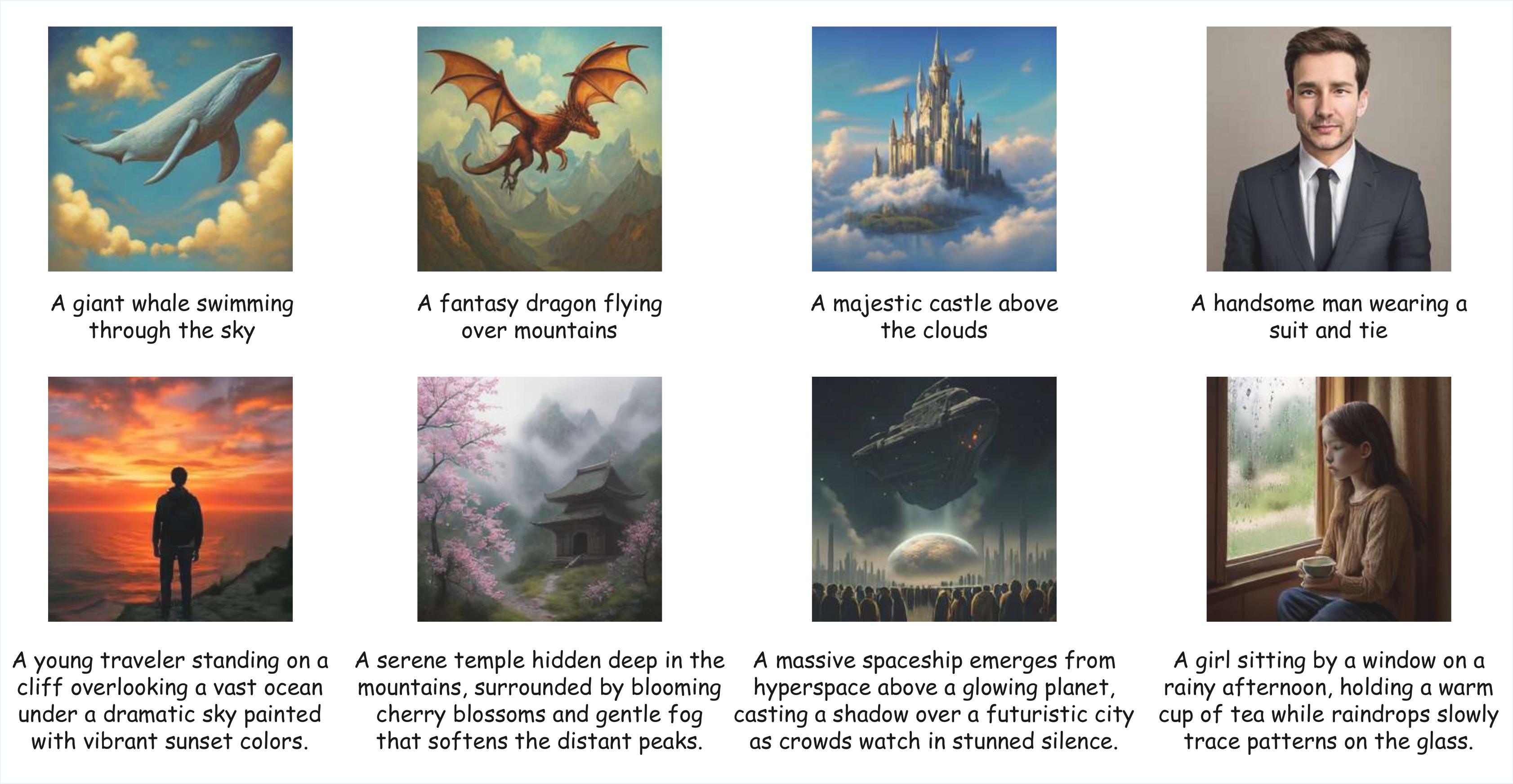}
    \caption{\textbf{Qualitative results of visual generation.}}
    \label{fig:vis_gen}
\end{figure}

We present additional qualitative results of visual generation in Fig.~\ref{fig:vis_gen}.
Our WinTok-based MLLM can generate high-quality and diverse images given either simple or complex text prompts.

\section{More Comparisons with UniTok}
\begin{figure}
    \centering
    \includegraphics[width=\textwidth]{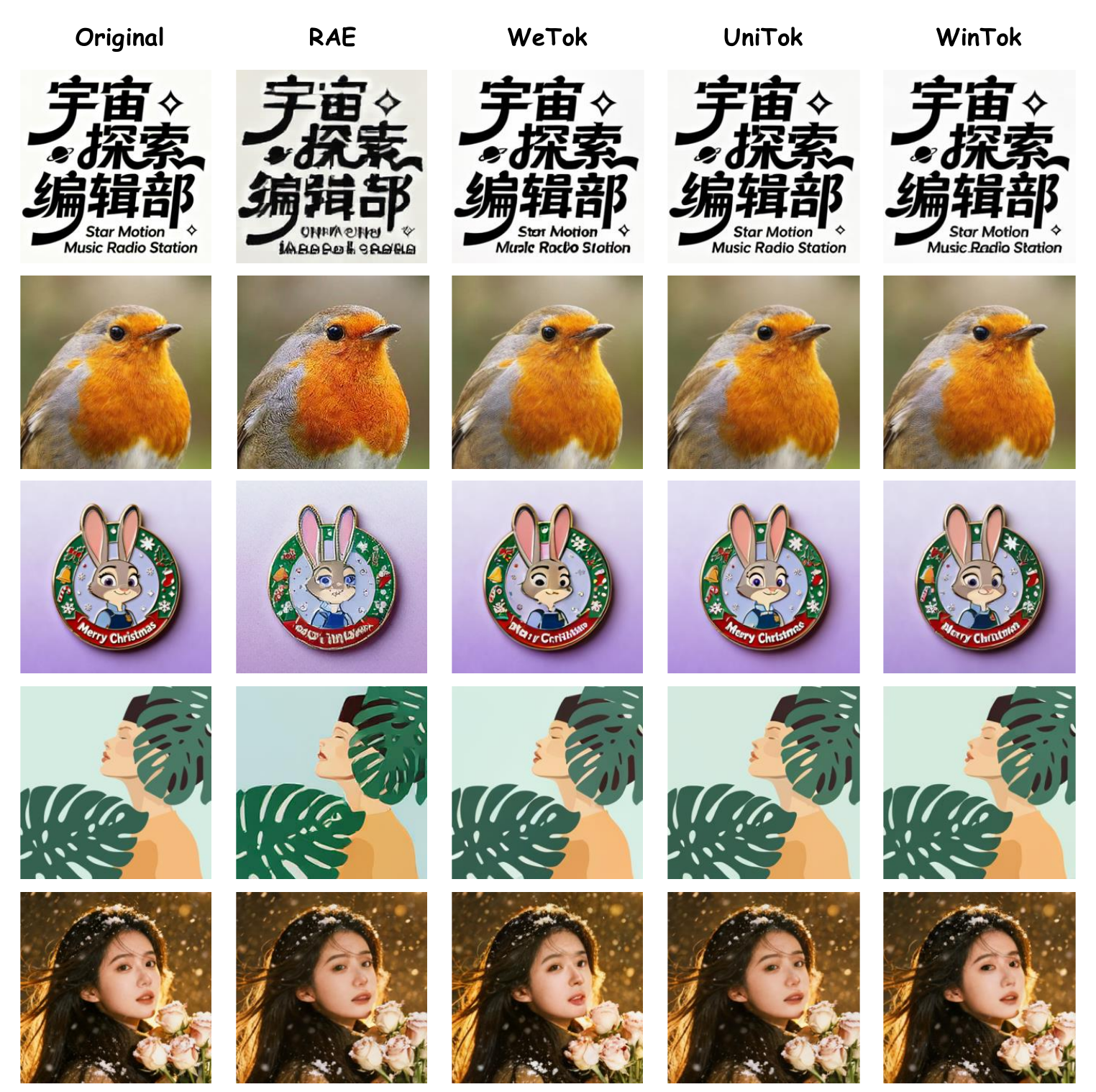}
    \caption{\textbf{Qualitative results of visual reconstruction.}
    All models are inferred at a resolution of 256\(\times\)256.
    }
    \label{fig:vis_2}
\end{figure}

\begin{figure}
    \centering
    \includegraphics[width=\textwidth]{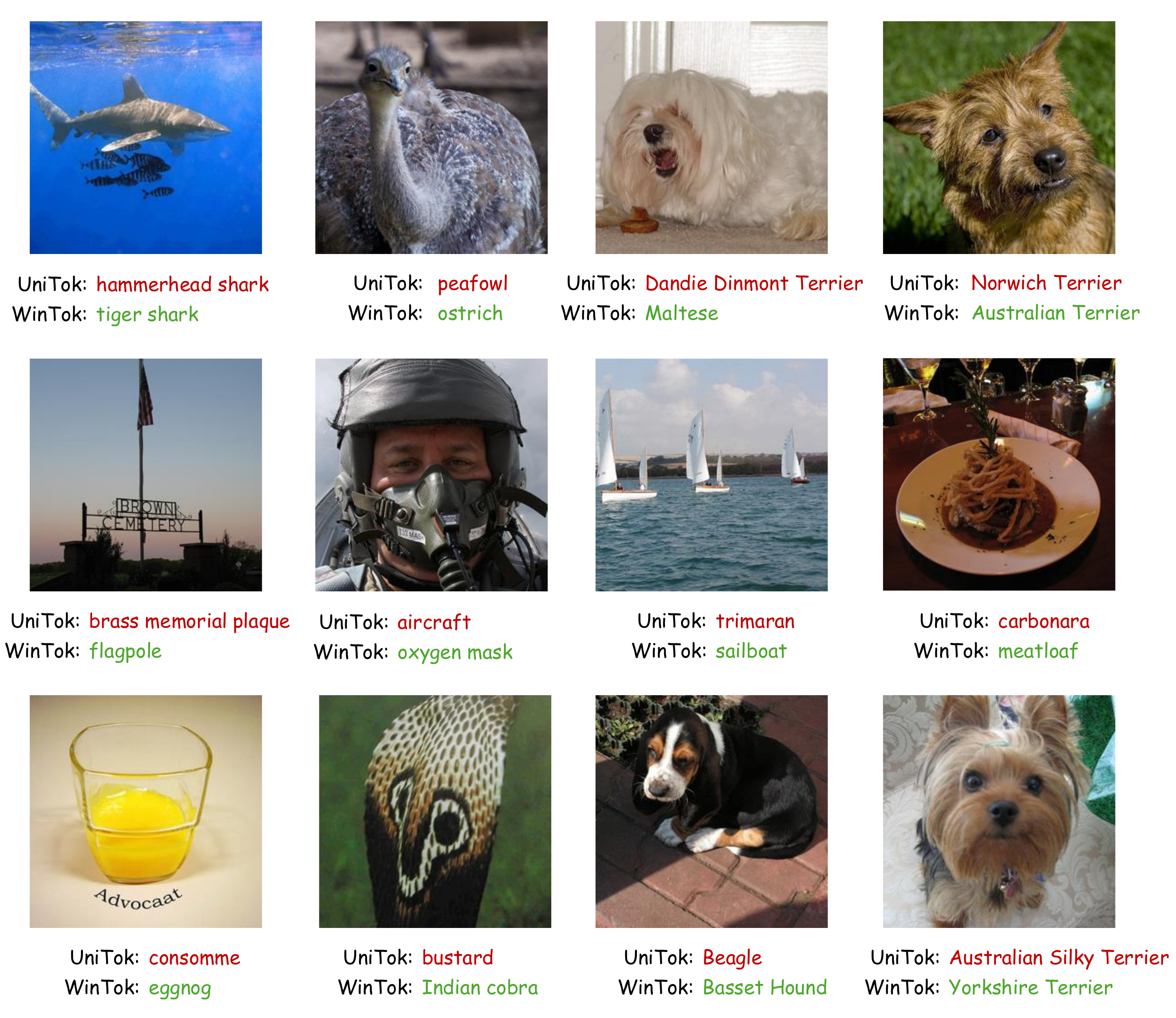}
    \caption{\textbf{Qualitative results of image classification.}
    All images are from ImageNet-1K validation set.
    }
    \label{fig:vis_cls}
\end{figure}

We provide more qualitative comparisons with the recent state-of-the-art unified visual tokenizer UniTok \cite{ma2025unitok}.
As for visual reconstruction, Fig.~\ref{fig:vis_2} shows that our WinTok demonstrates competitive reconstruction quality
compared to UniTok, while using significantly less training data (ImageNet-1K vs. DataComp-1B).
In terms of image classification, Fig.~\ref{fig:vis_cls} illustrates that our WinTok
can accurately classify various challenging images from ImageNet-1K validation set,
surpassing UniTok in recognition performance.




\clearpage